\documentclass[review]{elsarticle}

\usepackage{lineno,hyperref}
\usepackage{amssymb}

\usepackage{times}  % DO NOT CHANGE THIS
\usepackage{helvet} % DO NOT CHANGE THIS
\usepackage{courier}  % DO NOT CHANGE THIS
\usepackage{graphicx}  % DO NOT CHANGE THIS{\tiny }
\usepackage[namelimits]{amsmath} %数学公式
\usepackage{amsfonts}            %数学字体
\usepackage{mathrsfs}            %数学花体
\usepackage{color}
\usepackage{subfigure}

\usepackage{floatrow}
\newfloatcommand{capbtabbox}{table}[][\FBwidth]
\usepackage{float}

\journal{Applied Soft Computing}

\begin{document}

\begin{frontmatter}

\title{{Multi-fold Correlation Attention Network for Predicting Traffic Speeds with Heterogeneous Frequency
}}
%\thanks{Supported by organization x.}
%\tnotetext[mytitlenote]{This is a footnote associated with the title.}

%%% Group authors per affiliation:


\author[mymainaddress]{Yidan Sun}
\ead{ysun014@e.ntu.edu.sg}

\author[mysecondaryaddress]{Guiyuan Jiang\corref{mycorrespondingauthor}}
\cortext[mycorrespondingauthor]{Corresponding author}
\ead{jguiyuan@outlook.com}

\author[mymainaddress]{Siew-Kei Lam}
\ead{siewkei\_lam@pmail.ntu.edu.sg}

\author[mymainaddress]{Peilan  He}
\ead{phe002@e.ntu.edu.sg}

\author[mymainaddress]{Fangxin Ning}
\ead{fangxin\_ning@ntu.edu.sg}

\address[mymainaddress]{School of Computer Science and Engineering, Nanyang Technological University, 639798 Singapore}
\address[mysecondaryaddress]{College of Information Science and Engineering, Ocean University of China, Qingdao, 266000, China}

\begin{abstract}
{\color{black}
Short-term traffic prediction (e.g., less than 15 minutes) is challenging due to severe fluctuations of traffic data caused by dynamic traffic conditions and uncertainties (e.g., in data acquisition, driver behaviors, etc.).
Substantial efforts have been undertaken to incorporate spatiotemporal correlations for improving traffic prediction accuracy. 
In this paper, we demonstrate that closely located road segments exhibit diverse spatial correlations when characterized using different measurements, and considering these multi-fold correlations can improve prediction performance.
We propose new measurements to model multiple spatial correlations among traffic data. We develop a Multi-fold Correlation Attention Network (MCAN) that achieves accurate prediction by capturing multi-fold spatial correlation and multi-fold temporal correlations, and incorporating traffic data of heterogeneous sampling frequencies. 
The effectiveness of MCAN has been extensively evaluated on two real-world datasets in terms of overall performance, ablation study, sensitivity analysis, and case study, by comparing with several state-of-the-art methods. 
The results show that MCAN outperforms
the best baseline with a reduction in mean absolute error (MAE) by 13\% on Singapore dataset and 11\% on Beijing dataset.

}

\end{abstract}


\end{frontmatter}

%\linenumbers

\section{Introduction}
\label{Introduction}
Short-term traffic speed prediction (STTSP), which predicts the future traffic speeds in the next few minutes or even hours, is a core component in Intelligent Transportation Systems (ITS). The prediction accuracy critically impacts the performance of various applications, such as route planning, dynamic traffic management, and location-based service \cite{wang2014real}.
In the STTSP problem, the road network is typically organized as a graph, where the road segments are usually modeled as graph nodes while the graph links indicate the relationship between adjacent road segments.
{\color{black}
STTSP problem is challenging as the prediction accuracy is not only affected by environmental and periodical factors such as weather conditions (e.g., a strong storm), time of day (e.g., rush hours) and holidays,
but also by the complex dynamic spatiotemporal correlations among the traffic data.
A road segment's traffic speed series usually follows periodic patterns (e.g. daily, weekly periodicity) which change over time and vary geographically \cite{sun2019bus}. Also, the spatial correlation between different road segments' speed series is highly dynamic.
Compared with long-term traffic prediction, the STTSP problem is more challenging due to severe fluctuations of traffic data caused by dynamic traffic conditions and uncertainties (e.g., in data acquisition, driver behaviors, etc.).
}

The STTSP problem has been extensively studied, and various prediction models were proposed. It is well recognized that the traffic states in nearby regions are correlated with each other, which is also known as spatial correlations/dependencies. Many efforts have been devoted to incorporating spatiotemporal correlations to improve traffic prediction accuracy. 
In the early years, many classical methods (e.g., probabilistic graphical models \cite{koller2009probabilistic}, latent space model \cite{deng2016latent}) were developed for modeling pairwise correlations. However, they either use simple linear models which cannot model complex realistic scenarios or incur extremely high computational costs due to the large number of parameters.
{\color{black}
Recently, deep learning-based methods have been widely used for traffic prediction problems, due to their ability to model non-linear and non-stationary behaviors in traffic data. Many novel deep learning techniques (e.g., convolution neural network (CNN), graph convolutional network (GCN)) have been applied to traffic speed/flow prediction over road network structure for capturing spatial correlations \cite{chen2019gated,guo2020optimized,han2021dynamic,guo2021hierarchical}. 
To overcome the challenge of handling varying number of neighboring roads and connections in road networks, the work in \cite{chen2019gated} and \cite{guo2020optimized} utilized GCN combined with RNN to jointly capture spatial and temporal correlations. The difference is that, \cite{chen2019gated} applied a diffusion convolution layer directly on the road graph, while \cite{guo2020optimized} learned from optimized graphs via an updating strategy during the training phase. 
A dynamic GCN is proposed for traffic prediction \cite{han2021dynamic} by constructing dynamic graphs, which reflect time-specific spatial dependencies of road segments. The local neighborhood information can then be aggregated by dynamic GCN and propagated over the dynamic adjacent matrices. It is highlighted in \cite{guo2021hierarchical} that traffic systems are hierarchical structures consisting of micro layers (road networks) and macro layers (region networks); however, current GCN is only applied on the micro graph. Therefore, they applied a novel GCN-based traffic prediction method to capture features from both layers and the dynamic interaction between them. Although various methods have been developed to model spatial correlations, they are solely based on a single measurement to characterize the correlations (typically the traffic speed values). In this paper, we will demonstrate that spatial correlations can be characterized through multiple measurements such as the speed value, changing trends in speed values, and deviations from normal speed values. We will also explore and incorporate the effectiveness of multi-fold temporal correlations.
}

{\color{black}
Next, we show that closely located road segments exhibit a diverse set of spatial correlations when they are characterized using different measurements. These correlations alternatively dominate the spatial correlation, i.e., produce the highest correlation score.} 
For example, two road segments $r_a$ and $r_b$ may both observe similar speed values (e.g. $v_a$=20m/s, $v_b$=17m/s) at different time (e.g. 7:00 am and 11:50 am), indicating the same traffic situation at the two time instances. 
However, roads $r_a$ and $r_b$ may exhibit different changing trends at those time instances. For example, the traffic speed on both $r_a$ and $r_b$ tend to decelerate at 7:00am, while at 11:50am, the traffic speed at $r_a$ tends to decrease but the traffic speed at $r_b$ increases. As such, the spatial correlation measured based on changing trends is different from that based on traffic speed. 
There also exist many other measurements to characterize the spatial correlations, such as how much the current traffic speed deviates from its historical average speed. 
{\color{black} Specifically, the multi-fold spatial correlations studied in this paper refer to the multiple spatial correlations between two traffic speed series calculated based on multiple measurements. The measurements include not only the speed value but also the speed changing trend and deviation of current speed from the historical average.
A more detailed comparison of different spatial correlations will be illustrated later in Figure \ref{FigTrendDevCorr}. 
We will show that dynamically capturing and incorporating such multi-fold spatial correlations into the prediction model is vital for achieving higher prediction accuracy.
}
Adopting multi-fold temporal correlations for traffic predictions has been previously explored. For example, the daily and weekly periodicity are jointly considered for traffic prediction on grid network \cite{yao2019revisiting} and multi-resolution temporal correlations are adopted for traffic flow prediction on graph network. However, these methods fail to simultaneously model the multi-fold spatial correlations of traffic data.
In addition, actual traffic data are heterogeneous due to diverse road types (e.g. primary road, highway), travel demand, as well as the environmental and periodic factors (weather, time-of-day, events). 
As such, it is common for different road segments to exhibit traffic speed series with diverse frequencies. 
{\color{black}
In this paper, we introduce heterogeneous frequency in traffic speed data by considering different sampling rates for different road segments in the entire traffic network. 
For example, road segment $r_a$ has a speed observation per 5 minutes while $r_b$ generates a speed observation per 10 minutes.
}
Also, the generated speed series sometimes fluctuate notably due to missing values. 
Accurate traffic speed prediction on the graph road network with {\color{black}heterogeneous time-frequency traffic data} still remains an unsolved problem.

In this paper, we propose a Multi-fold Correlation Attention Network (MCAN) for traffic speed prediction that takes into consideration the heterogeneous frequency of traffic speed series. MCAN jointly takes into account multi-fold spatial correlations and multi-fold temporal correlations to provide discriminating features for improving the accuracy of traffic speed prediction. The major contributions of this paper are summarized as follows.

1) We propose alternative measurements to characterize the spatiotemporal correlations of the traffic data, i.e. how are they correlated with each other in terms of speed changing trend and the deviation from historical average speed. We demonstrate that the correlation patterns based on different measurements behave differently under various traffic situations, and they alternatively dominate the spatial correlation (i.e. produce the highest correlation score).

2) We propose a Heterogeneous Spatial Correlation (HSC) model to capture the spatial correlation with a specific measurement. 
The HSC model is able to deal with traffic data of heterogeneous sampling frequency.

3) We propose a novel MCAN framework to accurately predict traffic speed by simultaneously incorporating multi-fold spatial correlations, multi-fold temporal correlations as well as multiple contextual factors. To the best of our knowledge, our work is the first attempt to explicitly explore the multi-fold correlations for traffic prediction.
MCAN relies on the HSC model to explore multi-fold spatial correlations (based on the traffic speed, the speed changing trend and the deviation from historical average speed) and leverage upon the Long Short Term Memory (LSTM) model to capture multi-fold temporal correlations (short-term dependency, daily and weekly periodicity). The learned multi-fold spatiotemporal correlations together with contextual factors are fused with an attention mechanism to make the final predictions. 

4) We conduct extensive experiments on real-world traffic datasets. The results demonstrate that the proposed MCAN model outperforms the state-of-the-art baselines, and significant prediction accuracy improvement can be achieved by incorporating multi-fold correlations.

{\color{black}
The rest of the paper is organized as follows. Section 2 reviews the related works and highlights the similarity and differences between this work and the existing ones. Section 3 presents the necessary preliminaries, which include notations, problem formulation, and demonstration of multi-fold spatial correlations. Section 4 introduces our proposed MCAN model as well as the details of each component in MCAN. Section 5 evaluates the performance of the proposed methods using two real-world trafﬁc datasets. Section 6 concludes the paper and discusses potential directions of future work. 
}

\section{Related Work}

Different taxonomies have been utilized to review the existing methods, i.e.,  parametric approach and nonparametric approach \cite{van2012short}; statistically based methods, machine learning methods, and hybrid methods \cite{alsolami2020hybrid}.
Since most of the recent works rely on deep learning-based methods (including our work), in this paper, we divided related works into two categories \cite{du2019deep}: \textit{traditional shallow prediction models} and \textit{deep learning-based prediction models}. 

1) \textit{Traditional shallow prediction models} indicate traditional methods that are relatively simple and shallow in the model structure.
Reported methods in this category include Time Series Analysis (e.g. Auto-Regressive Integrated Moving Average (ARIMA) \cite{duan2016starima}), Kalman Filters (KF) \cite{mir2016adaptive}, K-nearest neighbor (KNN) based methods \cite{oh2015improvement,luo2019spatiotemporal}, Support Vector Machine (SVM) \cite{feng2018adaptive}, Markov models \cite{hu2016crowdsourcing}, Tensor decomposition based methods \cite{deng2016latent}, Gaussian Processe (GP) approaches \cite{schulz2018tutorial}, Random Forest (RF) methods \cite{yang2017ensemble}, and shallow neural network (NN) \cite{khosravi2011genetic}.

The above methods are often considered shallow prediction methods, as their model structures are simple and typically requires the developer to have some prior knowledge regarding the data distribution (e.g., ARIMA, KF, KNN, Markov, GP) or to design effective features for training the prediction models (e.g., SVM, GP, RF, shallow NN). This is in contrast to the deep learning methods, where one can supply fairly raw formats of data into the learning system, and the system can automatically learn feature representations regarding local and global relationships or structures in the data.
In addition, existing works have shown that the traffic speed series have inherent \textit{temporal patterns} (e.g., daily and weekly periodicity) \cite{yu2017deep} %chen2013road
and spatial dependencies (i.e., traffic states in nearby regions are correlated with each other) \cite{wang2016traffic}. However, the traditional shallow methods have difficulty in incorporating such temporal and/or spatial correlations, which are critical for producing prominent performance. 
As such, the existing shallow prediction methods typically failed to produce satisfying results for the challenging short-term traffic prediction.

2) {\textit{Deep learning-based methods}} provide a promising way to capture nonlinear temporal and spatial correlations for traffic prediction.
Traffic speed typically repeats periodically \cite{hou2016repeatability}, meaning that the traffic speed at a certain period is similar to the same time period of the previous day or previous week (i.e., temporal correlation).
As such, many existing works relied on recurrent neural networks (e.g., LSTM) \cite{cui2016deep} to incorporate the effect of temporal correlation. 
In fact, severe fluctuations were observed from the daily and weekly repeatability patterns due to dynamic traffic situations and uncertainties. 
A periodically shifted attention mechanism is developed in \cite{yao2019revisiting} to handle the long-term periodic temporal shifting.
In addition, some works also adopted the core idea of multi-fold temporal correlations for traffic prediction problems. For example, the daily and weekly periodicity are jointly considered for traffic prediction on grid network \cite{yao2019revisiting}, and multi-resolution temporal correlations are adopted for traffic prediction on graph network \cite{fang2019gstnet}.
However, these works incorporate multiple temporal correlations but fail to simultaneously model the multi-fold spatial correlations of traffic data.

To take into account the spatial correlations, many Convolutional Neural Networks (CNN) based methods \cite{zhang2017deep,asadi2020spatio} have been developed for traffic prediction problems by aggregating correlated neighborhood information (nearby road segments or regions). 
However, the methods are restricted to grid networks or ring road networks where each road segment has a fixed number of upstream and downstream road segments. 
As graph structure provides a natural representation of the traffic network, graph embedding techniques were developed to capture local-spatial correlations by aggregating the neighborhood information of nearby regions \cite{dai2016discriminative}. 
Graph Convolutional Neural Network (GCNN) \cite{henaff2015deep} is an appealing choice for modeling spatial correlation among graph-structured traffic data and has been proven to be very efficient for short-term traffic prediction \cite{chen2019gated}.
Earlier studies \cite{yu2018spatio} 
%\cite{yu2018spatio,puy2017unifying}
on traffic prediction with GCNN typically consider only static spatial dependencies, and the dynamic spatial dependencies were considered in the subsequent works to take into account gradual structural evolution of the road network \cite{yao2018deep}.
Spectral-based GCNN methods explore an analogical convolution operator over non-Euclidean domains on the basis of the spectral graph theory \cite{yao2018deep},
while the spatial-based GCNN methods always analogize the convolutional strategy based on the local spatial filtering
\cite{chai2018bike}. %\cite{chai2018bike,li2018diffusion,hechtlinger2017generalization}.
The existing methods typically considered a single type of spatial correlation and neglect the fact that the spatial correlations among the locations/roadways are diverse depending on their geographic attributes and traffic patterns.

The core idea of multi-fold spatial correlations bears some similarity with the concept of multi-view spatial correlations \cite{geng2019spatiotemporal} or diverse spatial correlations \cite{pan2019urban}. 
The work in \cite{geng2019spatiotemporal} proposed a spatiotemporal multi-graph convolution network (ST-MGCN) for ride-hailing demand forecasting. 
ST-MGCN considered multi-view spatial correlations including 
not only the correlations among spatially adjacent regions but also the pair-wise correlations among distant regions using multi-graph convolution. 
The work in \cite{pan2019urban} proposed a deep-meta-learning-based model (entitled ST-MetaNet) for predicting traffic conditions at all locations of an urban area. 
ST-MetaNet incorporated diverse types of spatiotemporal correlations by considering two types of geo-graph attributes 1) node attributes: the surrounding environment of a location, namely, nearby points of interests (POIs) and the density of road networks (RNs); 2) edge attributes: the relationship between two nodes, such as the connectivity of roads and the distance between them.
Although the above works considered multiple types of spatial correlations, our multi-fold spatial correlations are inherently different from the existing ones. 
Specifically, the existing works considered the correlations between the target roadway/location and multiple geographic components (e.g., location, region, points of interest (POIs), roadways, etc.), which are of different types. The correlation strength between two geographic components is characterized using a single measurement (e.g., traffic speed, network connectivity, etc.). 
Based on the dynamic situations, their prediction models selectively pay different attention to different correlations, which correspond to different geographic components.
In this paper, we observe that the correlation between two roadway segments can be characterized by multiple measurements (e.g., traffic speed, speed changing trend, etc.), entitled as multi-fold correlations.
The correlations based on different measurements have different patterns, and alternatively show the strongest correlation scores across various traffic situations. This means that, incorporating the multi-fold correlations relying on multiple measurements can better characterize the relationship between two roadways. 
More detailed discussion regarding the multi-fold spatial correlations will be presented in Section \ref{MF_SCM}.

In addition, the existing traffic prediction methods have largely neglected the heterogeneous frequencies over the temporal dimension of traffic data. 
The work in \cite{ye2012short} is the first one that considered data recorded at irregular time intervals, which would allow transportation management systems to be able to predict traffic speed based on intermittent data sources. The study adapted 3 existing parametric prediction methods (i.e., Naive Method, Simple Exponential Smoothing Method, and Wright's Modification of Holt's Method) to enable them to process irregular data through some transformations (i.e., treating a exponentially smoothing as exponentially weighting process and eliminating the weights corresponding to missing data interval). 
However, this method needs to customize the transformation operations for dealing with irregular time intervals for each of the three prediction models separately.
Also, it failed to incorporate spatial and temporal correlations among roadways, which are critical for prediction performance. 
Another work \cite{peng2019frequency} also considered heterogeneous frequency in traffic flow prediction, by dynamically filtering the inputs through Discrete Fourier Transform (DFT) on traffic flow data. 
With the filtered tensors, a 3D convolutional network is designed to extract the spatiotemporal features automatically, and several kernels with various sizes on the temporal dimension are employed to model the temporal correlations with multi-scale frequencies.
However, this method is limited to apply to grid-based traffic data and is only evaluated for coarse-grained data in both spatial (e.g., 1km $\times$ 1km) and temporal (15 minutes to 1 hour) dimensions. 
Different from the existing works, in this paper, we aim to develop an efficient method to process {\color{black}heterogeneous time-frequency traffic data} on fine-grained urban transport network while sufficiently incorporating spatiotemporal correlations.

\section{Preliminary}

\subsection{Notations}

\begin{table*}[htbp]
	\centering  
	\fontsize{7pt}{8pt} \selectfont
	\caption{Notations.}
	\label{NotationsTable}
	\begin{tabular}{l | l }
		\hline
		Notation & Description\\ 
		\hline
		$G=(V,E)$ & Traffic network.\\
		$r_i$ & The $i$th road segment.\\
		$\textbf{\^{y}}_{i} \in \mathbb{R}^{K_{i}}$ & Traffic speed series of $r_i$, $K_{i}$ is the length $\textbf{\^{y}}_{i}$.\\
		 $\textbf{\^{Y}} = \cup \textbf{\^{y}}_{i}$ & A set of traffic speed series of all roads. \\
		 $\bar{y}_{i}^{t}$ & The historical average traffic speed of all days at time interval t.\\
		 $\textbf{x}^{tr}_i$ & The speed changing trend vector of $r_i$.\\
		 $\textbf{X}^{tr}$ & A set of $\textbf{x}^{tr}_i$ of all road segments.\\
		 $\textbf{x}^{de}_i$ & The speed deviation vector of $r_i$.\\
		 $\textbf{X}^{de}$ & A set of $\textbf{x}^{de}_i$ of all road segments.\\
		 $\textbf{Y}\in \mathbb{R}^{N\times H}$ & The predicted traffic speed matrix on the entire network over next H time slices.\\
		 $T_{i}^{d}$ & The number of speed observations of $r_i$ in one day. \\
		 $y^{tr}_{i,t}$ & The speed changing trend of $r_i$ at time $t$. \\
		 $y^{de}_{i,t}$ & The speed deviation of $r_i$ at time $t$. \\
		 $\mathbf{e}_{i}^{t}\in \mathbb{R}^{c}$ & The output embedded vector of embedding component, $c$ is the length of vector.\\
		 $f_{CPA}$ & The function represents learnable Chebyshev Polynomial Approximation for replace operation.\\
		 $\{v_1,\cdots,v_{K_{em}}\}$ & The $K_{em}$ learnable coefficients corresponding to the polynomials.\\
		 $M\in \mathbb{R}^{c\times c}$ & A matrix with $c\times c$ learnable parameters to measure the correlation between embedded vectors.\\
		 $\sigma(\cdot)$ & The Sigmiod function.\\
		 $N_h(r_i)$ & The neighborhood of 1-hop to $h$-hop neighbor of $r_{i}$.\\
		 $f(\cdot)$ & The function repesents learnable Chebyshev Polynomial Approximation for GCN. \\
		 $\{z_1,\cdots,z_{K_{GCN}}\}$ & The $K_{GCN}$ learnable coefficients corresponding to the polynomials.\\
		 $\mathbf{h}^{tr}$, $\mathbf{h}^{de}$, $h^{S}$ & The three output vectors of HSC module. \\
		 $X^{recent}$, $X^{daily}$, $X^{weekly}$ & The three constructed input feature matrices of MTC module.\\
		 $X^R$, $X^E$ & The two constructed input feature matrices of contextual factor module.\\
		 $\{h_{1}(x),\cdots,h_{K_{em}(x)}\}$ & The $K_{em}$ truncated polynomials of $f_{CPA}$.\\
		 $\{h_{1}(x),\cdots,h_{K_{GCN}}(x)\}$ & The $K_{GCN}$ truncated polynomials of $f$.\\
		 $\alpha$, $\beta$ & hyperparameters in loss function, which control the weight of last two terms.\\
		 $d_{l}^{H}$, $d_{f}^{H}$ & The dimension of hidden layers of LSTM and FCN in Contextual Factors module.\\
		 $d_{l}^{T}$ & The dimension of hidden layers of LTSM in MSC module.\\
		 $d_{l}^{C}$, $d_{f}^{C}$ & The dimension of hidden layers of LSTM and FCN in HSC module.\\
		 $d_{o}$ & The length of output dimension of HSC, MSC, and contextual factor module.\\
		 $d_{a}$ & The dimension of hidden layers of attention layers.\\
		\hline %\cline{2-10}  
	\end{tabular} 
\end{table*}

\begin{figure}
	\centering
	\includegraphics[width=0.5\textwidth]{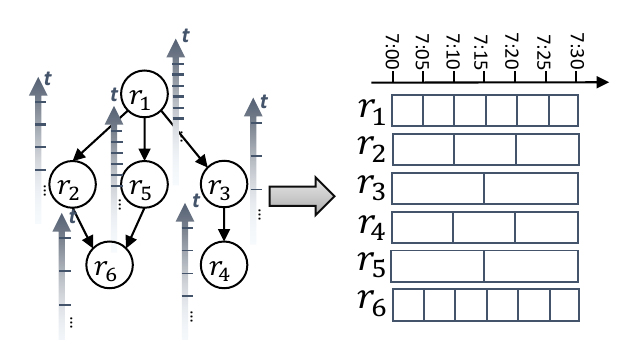}
	\caption{Traffic speed series with heterogeneous sampling frequency, where $r_1$ and $r_6$ has time interval of $5$ mins, $r_2$ and $r_4$ have time interval of $10$ mins, $r_3$ and $r_5$ has time interval of $15$ mins.} 
	\label{fig3}
\end{figure}

The traffic network can be represented as a graph $G=(V,E)$, where $V=\{$ $r_1$, $r_2$, $\cdots$, $r_N\}$ is a set of $N$ nodes (each represents a road segment) and $E$ is a set of edges. {\color{black} The graph is a directed graph and a two-way road segment is represented via two nodes. Both nodes and edges are unweighted.} 
Each node of $G$ generates a traffic speed series. 
In the same observation time span, the number of traffic speed values for different road segments are different due to heterogeneous sampling frequency.
As shown in Figure \ref{fig3}, roads $r_1$ and $r_6$ have {\color{black}time intervals} of $5$ mins between consecutive observations, $r_2$ and $r_4$ have {\color{black} time intervals} of $10$ mins, $r_3$ and $r_5$ has {\color{black}time intervals} of $15$ mins.
A vector $\textbf{\^{y}}_{i} \in \mathbb{R}^{K_{i}}$ is used to denote the traffic speed series of road segment $r_i$ over the entire time span, where $K_{i}$ is the number of speed values (or observation time intervals) in the observation time span. $\textbf{\^{Y}} = \cup \textbf{\^{y}}_{i}$ denotes the traffic speed series of all roads.

For each road $r_i$, a speed changing trend vector $\textbf{x}^{tr}_i$ is calculated, and $\textbf{X}^{tr} = \cup \textbf{x}^{tr}_i$.
{\color{black}Specifically, \textbf{$\textbf{x}^{tr}_i = \langle \Delta_{1}{\hat{y}_i}^{2}, \Delta_{1}{\hat{y}_i}^{3}, \cdots, \Delta_{1}{\hat{y}_i}^{K_i} \rangle$} where $\Delta_{1}{\hat{y}_i}^{t} = {\hat{y}_i}^{t} - {\hat{y}_i}^{t-1}$, indicating the changing trend of the traffic speed at time interval ($t$).}
{\color{black} 
Similarly, we calculate a speed deviation vector $\textbf{x}^{de}_i$, and $\textbf{X}^{de} = \cup \textbf{x}^{de}_i$.
Specifically, $\textbf{x}^{de}_i$ = $\langle \Delta_{2}{\hat{y}_i}^{1}$, $\cdots$ , $\Delta_{2} {\hat{y}_i}^{K_i} \rangle$, where ($\Delta_{2} {\hat{y}_i}^{t} = {\hat{y}_i}^{t} - \bar{y}_i^{t}$), $\bar{y}_i^{t}$ is the historical average traffic speed of all days at time interval ($t$), and $\Delta_{2} {\hat{y}_i}^{t}$ indicates the deviation of current speed ${\hat{y}_i}^{t}$ from $\bar{y}_i^{t}$ at time interval ($t$).
}

\subsection{Traffic Speed Prediction}
Given a road graph $G=(V,E)$ and historical traffic speed of all nodes over past time slices $\textbf{\^{Y}}$, predict future traffic speed series $\textbf{Y}=(\textbf{{y}}_{1}, \textbf{{y}}_{2},\cdots, \textbf{{y}}_{N}) \in \mathbb{R}^{N\times H}$ of all road road segments on the whole traffic network over the next $H$ time slices, where $\textbf{{y}}_{i} = ({{y}_i}^{t+1}, {{y}_i}^{t+2},\cdots, {{y}_i}^{t+H})$ {\color{black}denotes the future traffic speeds of road segment $r_i$, ${{y}_i}^{t+1}$ denotes the predicted speed value at future time slice $t+1$.}.

\subsection{Demonstration of Multi-fold Spatial Correlations.}

\begin{figure}
	\centering
	\includegraphics[width=0.75\textwidth]{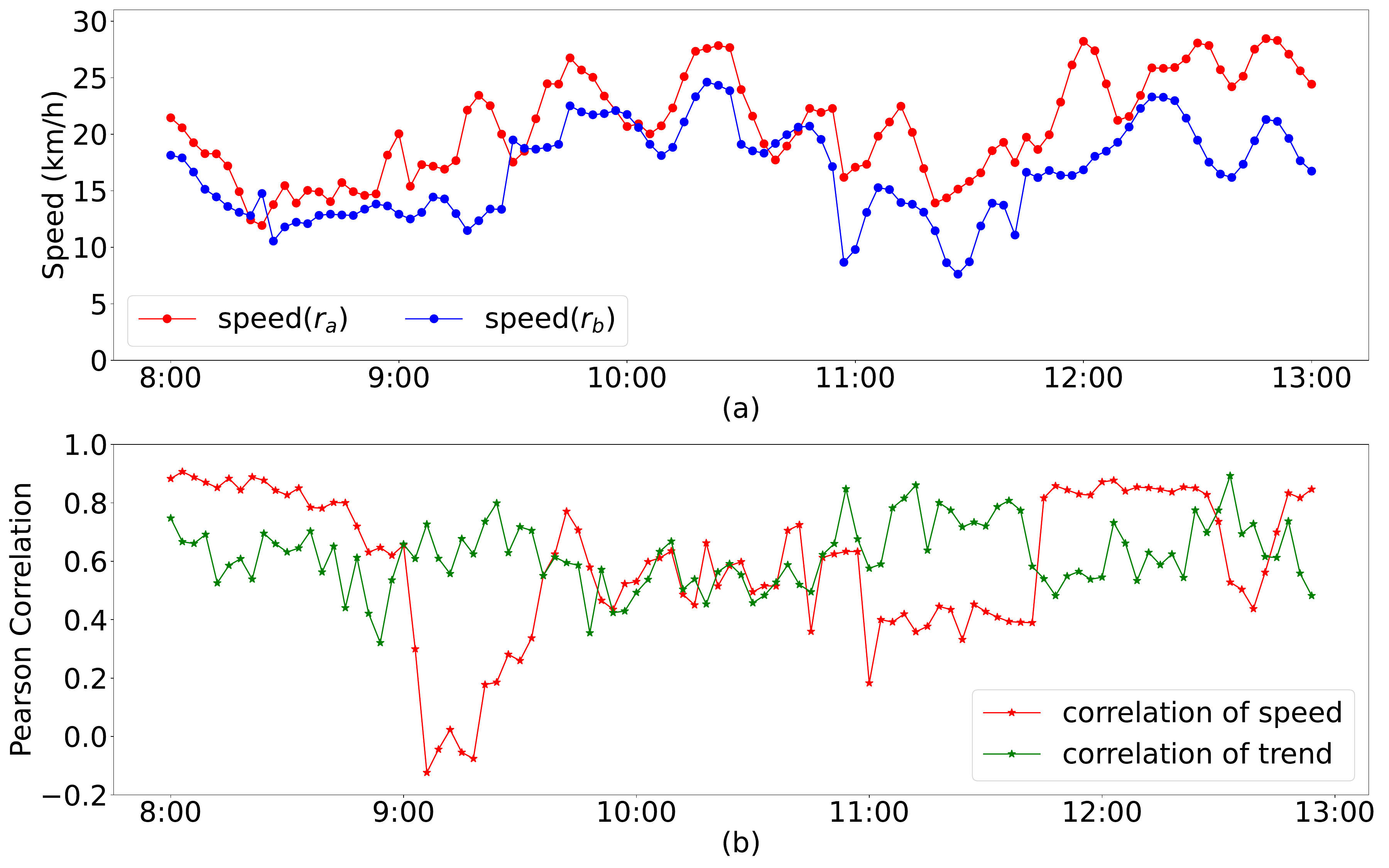}
	\caption{(a) traffic speeds of two road segments, $r_a$ and $r_b$, (b) correlations of $r_a$ and $r_b$ based on different speed measurements.} 
	\label{FigTrendDevCorr}
\end{figure}

Figure \ref{FigTrendDevCorr}(a) illustrates the traffic speeds of two segments $r_a$ and $r_b$ from 08:00 am to 13:00 pm. 
$r_a$ and $r_b$ tend to simultaneously increase (or decrease), which implies that $r_a$ and $r_b$ also have strong correlations in terms of speed changing trend. 
Figure \ref{FigTrendDevCorr}(b) compares the correlation scores obtained based on traffic speed and changing trend.
The correlation scores are calculated in the following way: for each time interval $t$, we calculate a traffic speed series of $r_i$ called ${\textbf{\^{o}}_i} = \langle {\hat{y}_i}^{t-d\ast T_{i}^{d}}, \cdots,  {\hat{y}_i}^{t-T_{i}^{d}}, {\hat{y}_i}^{t} \rangle$, where $T_i^d$ is the total number of speed observations of $r_i$ in one day, e.g., $T_i^d=288$ if $r_i$'s observation frequency is 5 minutes per record.
Thus, it consists of the traffic speed values of previous $d$ days at time interval $t$. 
Similarly, $\textbf{\^{o}}_j$ is calculated for $r_j$. Then the correlation between $r_i$ and $r_j$ with regards to time interval $t$ is obtained by calculating the Pearson Correlation between {$\textbf{\^{o}}_i$} and  {$\textbf{\^{o}}_j$}.
In the same way, the correlation curve based on changing trends is calculated and shown in Figure \ref{FigTrendDevCorr}(b).
The figure reflects that both the correlations between $r_i$ and $r_j$ change dynamically over time.
The two correlations have significantly different patterns, and they alternatively show the strongest correlation scores across various traffic situations. This indicates that it is important to consider multi-fold spatial correlations in order to capture the dynamic spatiotemporal characteristics of traffic data.

\section{Proposed Method}

\subsection{Main Framework}

\begin{figure}
	\centering
	\includegraphics[width=0.8\textwidth]{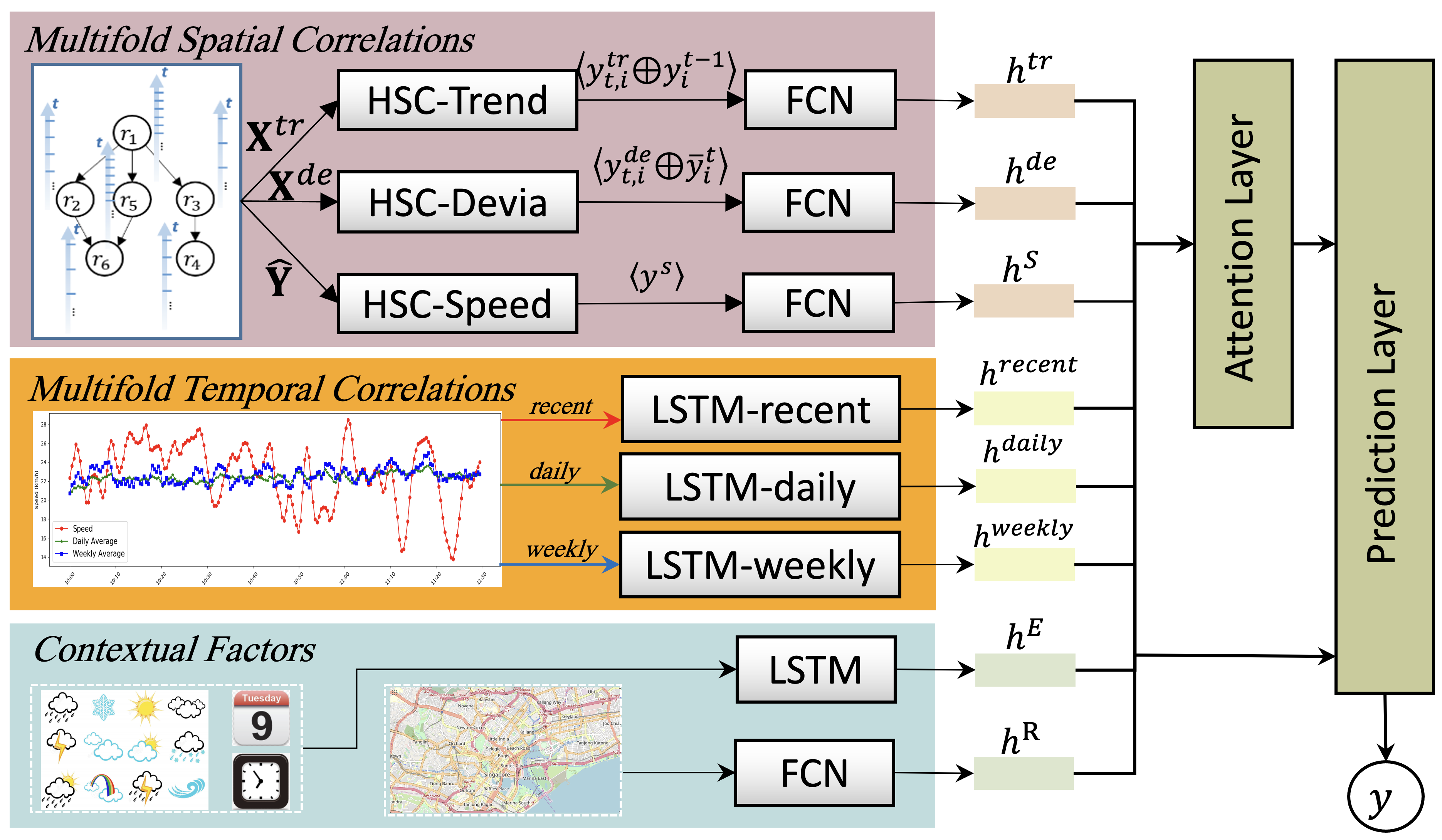}
	\caption{The framework of proposed MCAN Model.} 
	\label{Mainframework}
\end{figure}

{\color{black}
Figure \ref{Mainframework} presents the framework of our MCAN model, which consists of four major components. 
1) \textit{Multi-fold Spatial Correlation} module:
This module relies on an HSC model to learn multi-fold spatial correlations among the traffic data based on multiple measurements (including the traffic speed, the changing trend of traffic speed, and the deviation from historical average speed).
2) \textit{Multi-fold Temporal Correlation} module: 
This module leverages LSTM networks to learn multi-fold temporal correlations (namely, the short-term temporal dependency, as well as daily and weekly periodicity).
3) \textit{Contextual Factors} module:
This module leverages LSTM networks to learn the impacts of dynamic factors such as weather conditions and holidays and utilizes the FCN to learn the impacts of static road characteristics.
4) \textit{Attention Fusion} module: 
This component merges the learned multi-fold spatial correlations and the multi-fold temporal correlations together with contextual factors using an attention mechanism. Finally, the resultant fused features are fed into Fully Connected Neural Network (FCN) layers to produce the final predictions. 
}

\subsection{Multi-fold Spatial Correlation Module}\label{MF_SCM}

We first present the multi-fold spatial correlation (MSC) module, which explores the spatial correlations based on three different measurements of correlations, i.e. the traffic speed, the changing trend of the traffic speed (changing trend in short), and the deviation of the traffic speed from its historical average (deviation in short).
For changing trend, the MSC module relies on a Heterogeneous Spatial Correlation (HSC) model to learn a prediction $y^{tr}_{i} \in \mathbb{R}^{K_{i}}$ of changing trend for road segment $r_i$ based on $r_i$ and its neighbors' changing trend values. At the time $t$, $y^{tr}_{i,t}$ is concatenated with the traffic speed value at the previous time interval $y_i^{t-1}$, and then the concatenated information is fed into an FCN to learn a high-order feature representation. In the view of deviation, the HSC model learns a prediction of deviation $y^{de}_{i}$, which is concatenated with the average value of road $r_i$'s traffic speed and is further fed into an FCN to learn a high-order feature representation. For traffic speed, the output of the HSC model is directly fed into an FCN.

\begin{figure}
	\centering
	\includegraphics[width=\textwidth]{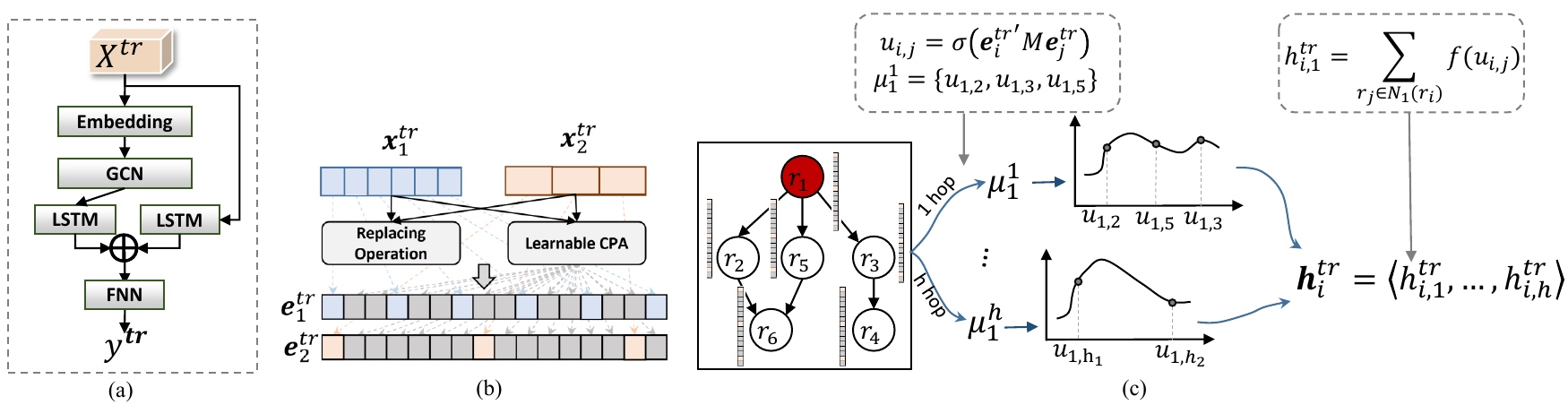}
	\caption{(a) the structure of the HSC model, (b) embedding component, (c) graph convolutional component (e.g. convolution operation process for node $r_1$ (red), $\mu^{1}_{1}$ is same as $N_{1}(r_1)$ which is the set of 1-hop neighbors of $r_1$).} 
	\label{correlation_model_details}
\end{figure}

We next explain how the HSC model capture spatial correlation in terms of changing trend, while the spatial correlations based on traffic speed or deviation can be processed in the same way. 
Figure \ref{correlation_model_details} shows the structure of the HSC model.
The HSC model consists of four main components: \textit{embedding} component, \textit{graph convolutional network (GCN)} component, \textit{LSTM} component and \textit{FCN} component. 
The HSC relies on the first three components to learn the neighborhood information and utilizes the LSTM component to learn a latent feature of each road segment's changing trend value.
Then the learned neighborhood information and the latent feature of the target road segment are concatenated and fed into an FCN layer to produce the output of the HSC model.

\subsubsection{Embedding Component} 
\begin{figure}
	\centering
	\includegraphics[width=0.6\textwidth]{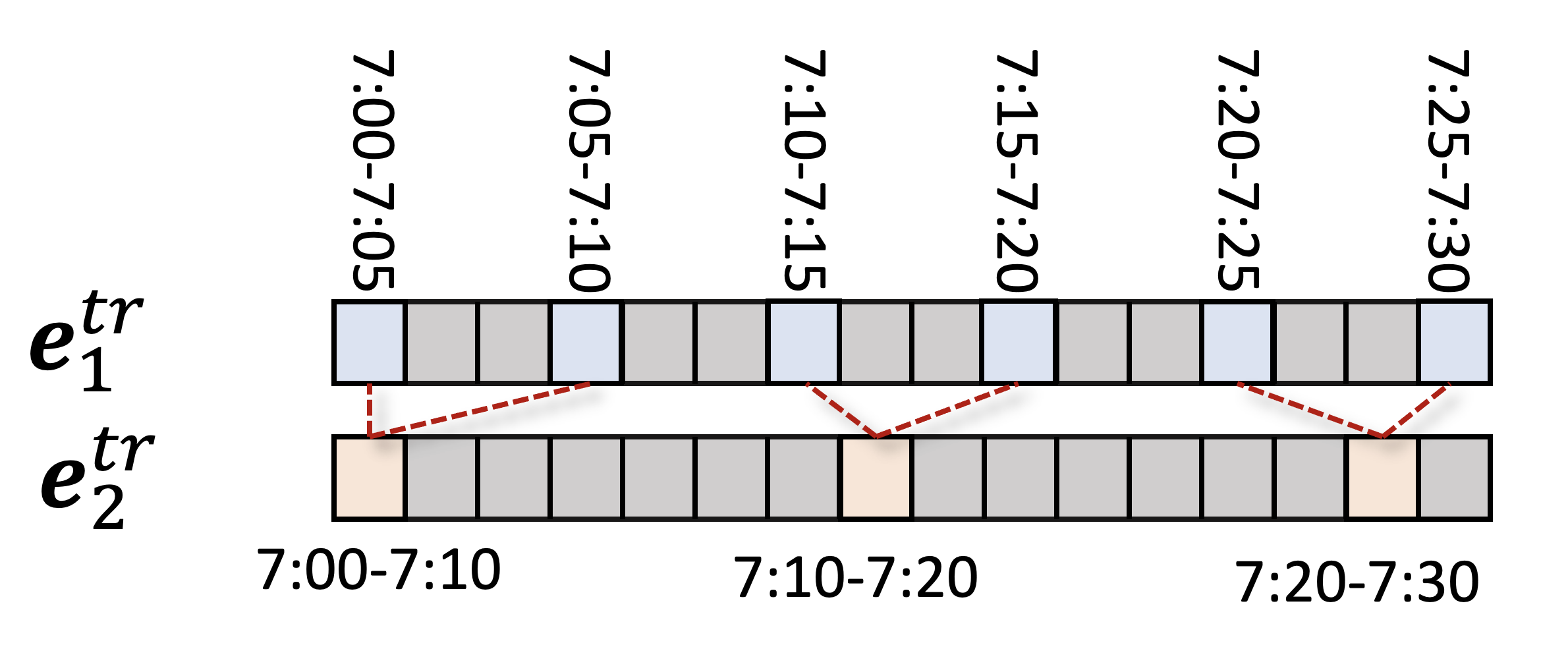}
	\caption{Illustration of keeping time consistency in embedded vectors.} 
	\label{time_consistency}
\end{figure}

This component takes $\textbf{X}^{tr}$ (the vector of changing trend of each road segment) as input where the length of different vectors can be different. The embedding component maps these vectors to the same feature space, and outputs a unified representation for various vectors with heterogeneous time granularity. 
For example, as shown in Figure \ref{correlation_model_details}(b) and Figure \ref{time_consistency}, $\textbf{x}^{tr}_1$ is a vector of $r_1$ with length of 6 (i.e. containing changing trend values for the past 30 minutes), and $\textbf{x}^{tr}_2$ is a vector of $r_2$ with length of 3. This means that  $r_1$ and  $r_2$ are heterogeneous in time granularity of the feature vectors. 
To deal with the heterogeneity, a replace operation combine with a {learnable} {C}hebyshev {P}olynomial {A}pproximation (CPA) \cite{chang2018structure} is used to embed the input vectors into vectors with unified length.
For an input vector $\textbf{x}^{tr}_i$, the embedding component distributes the elements of $\textbf{x}^{tr}_i$ into an embedded vector $\textbf{e}^{tr}_i \in \mathbb{R}^c$ as follows ($c$ is the length of the resultant vector),

\begin{equation} {\label{replacing}}
\textbf{e}^{tr}_i(j)=\left\{
\begin{array}{cl}
\textbf{x}^{tr}_i(m), & {\,j=m*(sn+1)}\\
f_{CPA}(j/len(\textbf{e}^{tr}_i)), & {otherwise}\\
\end{array} \right.             
\end{equation}
where $sn=\lfloor (c-len(\textbf{x}^{tr}_i))/(len(\textbf{x}^{tr}_i)-1) \rfloor$, $len(\textbf{x})$ is the length of vector \textbf{x}; {$m=0,1,2,...,len(\textbf{x}^{tr}_i)-1$}; 
$\textbf{e}^{tr}_i(j)$ is the $j$th element of the vector $\textbf{e}^{tr}_i$, $\textbf{x}^{tr}_i(m)$ is the $m$th element of the vector  $\textbf{x}^{tr}_i$;
the learnable CPA is denoted as $f_{CPA}(x) = \sum_{l=1}^{K_{em}}v_l\cdot h_l(x)$, where $h_l$ is the Chebyshev polynomial $h_l(x)$ of order $l$. $K_{em}$ is the number of the truncated polynomials, and $\{v_1,...,v_{K_{em}}\}$ are $K_{em}$ learnable coefficients corresponding to the polynomials $\{h_1(x),...,h_{K_{em}}(x)\}$.

The replacing operation ensures time consistency between embedded vectors, i.e. the elements with short temporal distance locate near each other in the embedded vectors. As shown in Figure \ref{time_consistency}, blue squares are from input vector $\textbf{x}^{tr}_1$ and pink squares are from input vector $\textbf{x}^{tr}_2$. The replacing operation guarantees that the blue and pink squares with close timestamps are placed close to each other in $\textbf{e}^{tr}_1$ and $\textbf{e}^{tr}_2$. 
For an input vector $\textbf{x}^{tr}_i$, after placing the elements of $\textbf{x}^{tr}_i$ to corresponding locations in $\textbf{e}^{tr}_i$, there are some elements in $\textbf{e}^{tr}_i$ with missing values. 
The CPA method is utilized to fill up the missing values within $\textbf{e}^{tr}_i$ by utilizing the Chebyshev polynomials approximation.
As shown in Figure \ref{correlation_model_details}(b), the coefficients inside the CPA function are learnable, which allows the model to automatically fit a distribution of input features.

\subsubsection{GCN Component} 
This component incorporates spatial information of local neighborhoods from 1-hop to $h$-hop neighbors. 
With the embedded feature vectors from the embedding component, a GCN \cite{chang2018structure} is used to aggregate the local neighborhood information from both the target $r_i$ and all its 1-$h$ hop neighbors. It was shown to be efficient for capturing spatial correlations while preserving the structure consistency.
As shown in Figure \ref{correlation_model_details}(c), the road segment $r_i$ and its neighbors each maintains an embedded vector of length $c$ obtained from the embedding component. 
Then the local information with regards to the neighborhood of 1-hop to $h$-hop neighbor is learned as: 
\begin{equation} {\label{GCN}}
\mu_i^h = \{u_{i,j}|r_j\in N_h(r_i)\},~
u_{i,j}=\sigma(({\textbf{e}^{tr}_i})' M\textbf{e}^{tr}_j),           
\end{equation}
\begin{equation} {\label{GCN_2}}
h_{i,h}^{tr}=\sum_{r_j \in N_h(r_i)}f(u_{i,j}),~
f(u_{i,j})=\sum_{l=1}^{K_{GCN}}z_l\cdot h_l(u_{i,j}),     
\end{equation}
where $\textbf{e}^{tr}_i,\textbf{e}^{tr}_j \in \mathbb{R}^c$ are embedded features of road $r_i$ and $r_j$, $({\textbf{e}^{tr}_i})'$ is the transposition of $\textbf{e}^{tr}_i$, $M \in \mathbb{R}^{c\times c}$ is a matrix with $c\times c$ learnable parameters to measure the correlation between embedding vectors, and $\sigma(\cdot)$ is the Sigmiod function, $K_{GCN}$ is the number of the truncated polynomials, and $\{z_1, ..., z_{K_{GCN}}\}$ are $K_{GCN}$ coefficients corresponding to the polynomials $\{h_1(x),...,h_{K_{GCN}}(x)\}$. 
The convolution filter is parameterized with numbered learnable parameters under the guidance of the function approximation theory \cite{chang2018structure}. 
The output of GCN for $r_i$ is $\textbf{h}^{tr}_i=\langle h_{i,1}^{tr},...,h_{i,h}^{tr} \rangle$, where $h_{i,h}^{tr}$ contains the neighborhood information of up to $r_i$'s $h$-hop neighbors, $N_h(r_i)$ represents the $h-$hop neighbors of $r_i$.

\subsubsection{LSTM Layer}
\label{LSTM}
The architecture of the LSTM cell can be described with the following equations:
\begin{equation}\label{LSTM_eq}
\begin{split}
\mathbf{i}_{tt} &= \sigma(\mathbf{W}_{ix}\mathbf{x}_{tt} + \mathbf{W}_{ih}\mathbf{h}_{tt-1} + \mathbf{b}_i), \\
\mathbf{f}_{tt} &= \sigma(\mathbf{W}_{fx}\mathbf{x}_{tt} + \mathbf{W}_{fh}\mathbf{h}_{tt-1} + \mathbf{b}_f), \\
\mathbf{o}_{tt} &= \sigma(\mathbf{W}_{ox}\mathbf{x}_{tt} + \mathbf{W}_{oh}\mathbf{h}_{tt-1} + \mathbf{b}_o), \\
\widetilde{\mathbf{C}_{tt}} &= \tanh(\mathbf{W}_{Cx}\mathbf{x}_{tt} + \mathbf{W}_{Ch}\mathbf{h}_{tt-1} + \mathbf{b}_C), \\
\mathbf{C}_{tt} &= \mathbf{i}_{tt}\ast \widetilde{\mathbf{C}_{tt}} + \mathbf{f}_{tt}\ast \mathbf{C}_{tt-1}, \\
\mathbf{h}_{tt} &= \mathbf{o}_{tt}\ast \tanh(\mathbf{C}_{tt}).
\end{split}
\end{equation}
where $tt$ stands for the $tt$-th time interval,  $\mathbf{i}_{tt}$, $\mathbf{f}_{tt}$, $\mathbf{o}_{tt}$ refer to the output of the input gate, forget gate and output gate respectively. 
$\mathbf{x}_{tt}$, $\mathbf{c}_{tt}$, $\mathbf{h}_{tt}$ 
are the input vector, state vector and hidden vector respectively, and 
$\mathbf{h}_{tt-1}$ is the former output of $\mathbf{h}_{tt}$.
$\widetilde{\mathbf{C}_{tt}}$ and $\mathbf{C}_{tt}$ are the input state and output state of the memory cell, and $\mathbf{C}_{tt-1}$ is the former state of $\mathbf{C}_{tt}$. 
$\sigma$ is a sigmoid function.
$\mathbf{W}_{ix}, \mathbf{W}_{fx}, \mathbf{W}_{ox}, \mathbf{W}_{Cx}$ are the weight matrices connecting $\mathbf{x}_{tt}$ to the three gates and the cell input, $\mathbf{W}_{ih}, \mathbf{W}_{fh}, \mathbf{W}_{oh}, \mathbf{W}_{Ch}$ are the weight matrices connecting $\mathbf{x}_{tt-1}$ to the three gates and the cell input, $\mathbf{b}_i$, $\mathbf{b}_f$, $\mathbf{b}_o$, $\mathbf{b}_C$ are the bias terms of the three gates and the cell gate.

Two separate LSTM components are used to learn the temporal dependency of the changing trend features: one is for the trend vector of $r_i$, and another is for $\textbf{h}^{tr}_i$ (the neighborhood information).
The outputs of the two LSTM components are denoted as $\mathbf{h}^{Neigh}$ and $\mathbf{h}^{Self}$ respectively. 
The two terms are concatenated and then fed into an FCN component to produce the output of the HSC model, i.e. prediction of changing trend $y_i^{tr}$ and of deviation $y_i^{de}$.

\subsection{Multi-fold Temporal Correlation Module}

{\color{black}

Temporal correlations have been widely considered by existing decomposition methods for time-series analysis. They typically split a series into a combination of level, trend, seasonality, and noise components. Such decomposition helps capture a variety of patterns exhibited by the time series data, where each component represents an underlying pattern category \cite{bokde2018analysis}. However, it requires predefined models to decompose the data, which may not correctly match the actual patterns of the data.
The core idea of the decomposition method has been considered by many deep learning models \cite{abdelraouf2021utilizing}.
In this paper, our MCAN model relies on the MTC module to capture and incorporate multiple temporal correlations, based on three different temporal dependencies, i.e., the recent dependency, the daily periodicity correlation, and the weekly periodicity correlation. 
The multi-fold temporal correlation is similar to decomposition methods in exploring the various temporal patterns of the time series data. The difference is that, decomposition methods explicitly decompose and model the various temporal patterns, while our MCAN relies on the MTC module to automatically learn the different temporal patterns from input data without decomposing the data explicitly. 
Specifically, the three LSTM modules share the same network structure but take different inputs. The LSTM structure has been introduced in the previous section, and the inputs to the three LSTM modules are as follows. 
}

\subsubsection{The Recent Correlation}
Intuitively, the formation and dispersion of traffic congestions are gradual, indicating that the traffic speed at previous time intervals (recent information) inevitably has influences on the future traffic speed.
Four kinds of recent information are utilized to capture the recent correlation:
1) {the traffic speeds of previous} $lr$ time slots, i.e. $\mathbf{rs} = <\hat{y}_{i}^{t-lr}, \hat{y}_{i}^{t-lr+1}, \ldots, \hat{y}_{i}^{t-1}>$, where $lr$ is the number of recent observations utilized.
2) the speed changing trend of previous $lr$ time interval, i.e. $\mathbf{rt} = <x^{tr}_{i,t-lr}, x^{tr}_{i,(t-lr+1)}, \ldots, x^{tr}_{i,(t-1)}>$.
%, where $T_d$ is the number of time intervals in each day.
3) the deviation of the traffic speeds of previous $lr$ time interval, i.e. $\mathbf{rd} = <x^{de}_{i,t-lr}, x^{de}_{i,(t-lr+1)}, \ldots, x^{de}_{i,(t-1)}>$.
4) {the average speeds at the previous $lr$ time slots, $\mathbf{ra} = <\bar{y}_i^{t-lr}, \bar{y}_i^{t-lr+1}, \ldots, \bar{y}_i^{t-1}>$, where $\bar{y}_i^{t-1}$ is the average traffic speed of all days at time interval $t$ for road $r_i$.}
The input to the LSTM-recent module is $\mathbf{X}^{recent}=<\mathbf{rs},\mathbf{rt},\mathbf{rd}, \mathbf{ra}>$.

\subsubsection{The Daily Periodicity Correlation}
Due to the regular daily journey routines taken by people, the traffic speed typically repeats periodically, e.g. the traffic speed at a certain period is similar to the same time period of the previous day \cite{hou2016repeatability}. 
We incorporate the periodicity information of a road segment at the time interval to improve prediction accuracy.
The daily periodicity information consists of three parts:
1) the traffic speeds of previous $ld$ days at the same time interval, i.e. $\mathbf{ds}$ = $<\hat{y}_{i}^{t-ld\ast T_{i}^{d}}$, $\hat{y}_{i}^{t-(ld-1)\ast T_{i}^{d}}$, $\ldots$, $\hat{y}_{i}^{t-T_{i}^{d}}>$, where $T_{i}^{d}$ is the number of time intervals in each day for $r_i$.
2) the changing trend of the traffic speeds of previous $ld$ days at the same time interval, i.e. $\mathbf{dt}$ = $<x^{tr}_{i,t-ld\ast T_{i}^{d}}$, $x^{tr}_{i,t-(ld-1)\ast T_{i}^{d}}$, $\ldots$, $x^{tr}_{i,t-T_{i}^{d}}>$.
3) the deviation of the traffic speeds of previous $ld$ days at the same time interval, i.e. $\mathbf{dd}$ = $<x^{de}_{i,t-ld\ast T_{i}^{d}}$, $x^{de}_{i,t-(ld-1)\ast T_{i}^{d}}$, $\ldots, x^{de}_{i,t-T_{i}^{d}}>$.
As such, the input to the LSTM-daily module is $\mathbf{X}^{daily}=<\mathbf{ds}, \mathbf{dt}, \mathbf{dd}>$.

\subsubsection{The Weekly Periodicity Correlation}

Traffic data also shows a strong weekly periodicity. 
The weekly periodicity information consists of three parts, based on traffic speed, changing trend and deviation respectively.
1) the traffic speeds of previous $lw$ weeks at the same time interval, i.e. $\mathbf{ws}$ = $<\hat{y}_{i}^{t-lw\ast T_{w}^{i}}$, $\hat{y}_{i}^{t-(lw-1)\ast T_{w}^{i}}$, $\ldots$, $\hat{y}_{i}^{t-T_{w}^{i}}>$, where $T_{w}^{i}$ is the number of time intervals in each week for $r_i$.
2) the changing trend of previous $lw$ weeks at the same time interval, i.e. $\mathbf{wt}$ = $<x^{tr}_{i,t-lw\ast T_{w}^{i}}$, $x^{tr}_{i,t-(lw-1)\ast T_{w}^{i}}$, $\ldots$, $x^{tr}_{i,t-T_{w}^{i}}>$.
3) the deviation of previous $lw$ weeks at the same time interval, i.e. $\mathbf{wd}$ = $<x^{de}_{i,t-lw\ast T_{w}^{i}}$, $x^{de}_{i,t-(lw-1)\ast T_{w}^{i}}$, $\ldots$, $x^{de}_{i,t-T_{i}^{w}}>$.
As such, the input to the LSTM-weekly module is $\mathbf{X}^{weekly}$ = $<\mathbf{ws}$, $\mathbf{wt}$, $\mathbf{wd}>$.

{\color{black}
\subsection{Contextual Factors Module}
\label{section:ContextFactor}
In addition to the multi-fold spatial/temporal correlations, \textit{contextual factors} are also considered, which include the road network characteristics and weather conditions. 
The contextual factors can be divided into two parts: static and dynamic factors.
The static factors are represented as a feature vector $\mathbf{X}^R$ to capture the road characteristics of a road segment. 
It includes the length of the road segment, road type (e.g., primary road, highway), number of lanes, and number of traffic lights.
The dynamic (time-dependent) factors are represented as a feature vector $\mathbf{X}^E$ to capture factors such as weather conditions and activity events. 
It includes the {weather condition} (one-hot encoding), {holidays}, {time interval} $t$ and the {day-of-week}. 
The static feature vector $\mathbf{X}^R$ is fed into an FCN layer, and the dynamic feature vector $\mathbf{X}^E$ is fed into an LSTM layer to learn high order representations.}

\subsection{Component Fusion with Attention for Prediction}
Then the multi-fold spatial correlations, multi-fold temporal correlations, as well as the contextual factors are fused together using an attention mechanism.
{\color{black}The resultant fused features are fed into a fully connected layer to produce the final predictions.}
The loss function contains three terms, i.e. mean square error of speed value, changing trend and deviation. 
$\alpha$ and $\beta$ are hyperparameters to control the weights of different terms. 
\begin{equation}\label{Loss}
L_{loss}= \sum_{i=1}^{N}({\parallel \hat{y_i}}-y_i\parallel^2 + \alpha\parallel{x}^{tr}_i-y^{tr}_i \parallel^2 + \beta\parallel{x}^{de}_i-y^{de}_i \parallel^2).
\end{equation}
The algorithm Adam is utilized for optimization. The training process repeats for 50 epochs. To prevent overfitting, the dropout mechanism
is applied to each hidden layer, where the rate of dropout is set to 0.5.

\section{Experiments}
\subsection{Settings}
\subsubsection{Dataset}
\label{section:exper:datasets}

\begin{figure}[tbp]
	\centering
	\subfigure{
		\label{fig_dataset_sg}
		\includegraphics[width=0.45\textwidth]{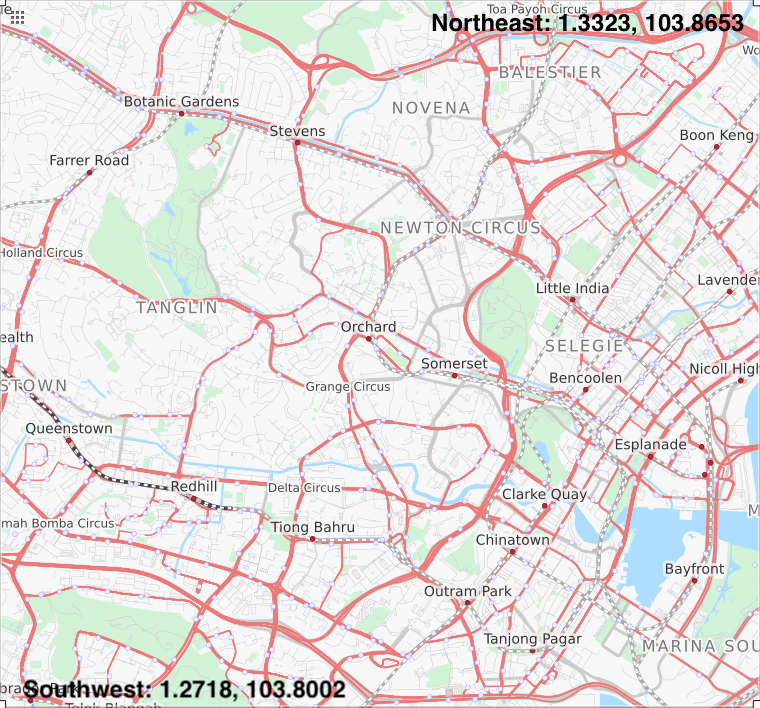}
		}
	\subfigure{
		\label{fig_dataset_bj}
		\includegraphics[width=0.48\textwidth]{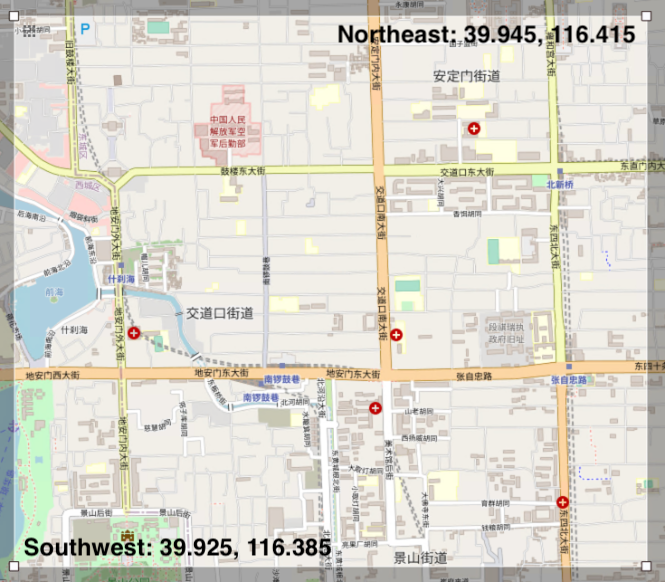}
		} 
	\caption{Illustration of two studied area, left: the downtown area in Singapore, right: the downtown area in Beijing.}
	\label{datasets}
\end{figure}

\noindent \emph{Road Network--Singapore:}
The first road network in Singapore is obtained from OpenStreetMap\footnote{https://www.openstreetmap.org/export}.
We choose a rectangle area in Downtown Singapore (Southwest: 1.2718, 103.8002; Northeast: 1.3323, 103.8653), which comprised of 974 road segments (we only consider road segments traversed by bus services) {as shown in Figure \ref{fig_dataset_sg}}.   
This is used to derive topological attributes of the road segments, which include: types of road segments (e.g. primary, residential, highway), number of lanes, length, availability of traffic signals at the end of road segments.

{\color{black}
\noindent \emph{Road Network--Beijing:}
The second road network in Beijing is obtained from \cite{liao2018deep}\footnote{https://github.com/JingqingZ/BaiduTraffic}.
We choose a rectangle area in Downtown Beijing (Southwest: 39.925, 116.385; Northeast: 39.945, 116.415), which is comprised of 523 road segments {as shown in Figure \ref{fig_dataset_bj}}. This is used to derive topological attributes of the road segments, which include the number of lanes, length, width, direction, the number of lanes, and speed class.
}

\begin{figure}[thb]
	\centering
	\includegraphics[width=0.75\textwidth]{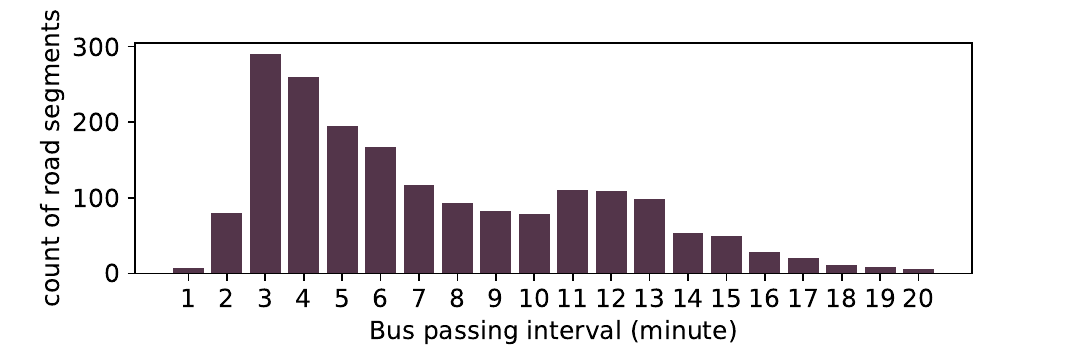} 
	\caption{Distribution of bus passing frequency of all road segments.}
	\label{road_frequency}
\end{figure}

\noindent\emph{Traffic Speed Data--Singapore:} 
The traffic speed data is calculated based on historical bus trajectories derived from bus arrival data \footnote{https://www.mytransport.sg/content/mytransport/home/dataMall.html} (which contains the bus locations of each bus service at each minute). 
All the bus locations of the same bus trip form a bus trajectory.
Each point of a trajectory contains the GPS location of the bus and the corresponding timestamp. We apply map-matching to project trajectories to road network for calculating the traffic speed of corresponding road segments. 
{\color{black}
Due to many factors such as device errors and network issues, there exist many missing values in the raw dataset. To mitigate this problem, we first fill up the missing values using spline interpolation \cite{mckinley1998cubic}.
Then, we remove the speed values that exceed the speed limits of the road segments.
}
In practice, different road segments have significantly different frequencies of bus passing. The time interval between consecutive buses ranges from 1 minute to 20 minutes. As such, different road segments have different sampling rates. 
The data frequency for a road segment is set based on the frequency of the bus services that traverse the road. For example, if bus 179 traverses road $r_i$ and its average time interval between consecutive bus arrivals is 7 minutes, then we calculate a traffic speed value in every 7 minutes for road $r_i$. If there are multiple bus services that cover the same road segment, the average time interval of all bus services is used.
Figure \ref{road_frequency} shows the distribution of bus passing frequency of all road segments. 
Bus traffic data from Aug. 01 to Nov. 30, 2018, are used in our experiment. Traffic data on the first 91 days are used as the training set, and the remaining are used as the testing set.
For the target road segment and all its 1-h hop neighbors, we consider traffic data of the past 1 hour, i.e., for a neighbor $r_j$ with time interval $T_j$, the past $60/T_j$ time intervals will be fed into the HSC model.

{\color{black}
\noindent\emph{Traffic Speed Data--Beijing:} 
The traffic speed data (Q-Traffic) is also obtained from \cite{liao2018deep}\footnote{https://github.com/JingqingZ/BaiduTraffic}. The time interval is 15 minutes for all road segments, the studied period from Apr. 01 to May. 31 2017. 
Traffic data on the first 45 days are used as training set, and the remaining are used as the testing set. 
}

\noindent\emph{Weather data:} 
Hourly-grained weather data are collected during the same time period of the bus speed data\footnote{https://www.timeanddate.com/weather/singapore/singapore},
{\color{black}
that include four parts: (1) weather categories (e.g., sunny, light rain, passing cloud, etc.), which consist of 52 categories, and we applied one-hot encoding to represent the categories; (2) temperature in real number; (3) wind speed in real number; (4) probability of precipitation. 
These four parts as well as holiday, time of day, and day of week are concatenated together as one feature vector into LSTM in Contextual Factors Module (as described in Section \ref{section:ContextFactor}).
}

\subsubsection{Baselines}
\label{BaselinesAndAblation}

We compare our method with the following baseline methods.
(1) {ARIMA}: Auto Regression Integrated Moving Average \cite{williams2003modeling}. 
(2) {LR}: Linear Regression\cite{gross2003linear}. 
(3) {SVR}: Support Vector Regression \cite{wu2004travel}. 
(4) {STGCN} \cite{yu2018spatio}: {\color{black} It applies pure convolution structure to simultaneously capture temporal dynamics and spatial dependencies in traffic data. They designed a module consisting of graph convolution layers to learn spatial correlations on traffic networks and gated convolutional layers to capture temporal correlations. The framework of STGCN consists of two blocks mentioned above and a fully-connected output layer at the end.}
(5) {BTSP} \cite{sun2019bus}: {\color{black} It utilized strucutre2vec technique to model spatial correlation within local neighborhoods. For global spatial correlation, they apply a cut-based partitioning method to cluster road segments with similar traffic patterns for training corresponding predictors. LSTMs are fed with carefully designed short- and long-term temporal features for modeling temporal dependencies. Finally, an attention layer fuses the above heterogeneous spatial and temporal features based on dynamic traffic situations for generating the final prediction result.}
(6) {ASTGCN} \cite{guo2019attention}: {\color{black} It includes three independent components of the same architecture to capture three temporal dependencies of the traffic: recent, daily- and weekly- periodic dependencies. In each component, there are two stacked ST blocks. In each ST block, 1) an attention mechanism is proposed to capture the importance of different locations and time slices, to enable the network to pay more attention to valuable information; 2) a graph convolution network combined with standard convolution layer aims to incorporate spatial and temporal features based on learned attention scores. Using the above ST blocks, three latent features are generated by the three components and are fused for final prediction.}
{\color{black}
(7) LSTM \cite{hochreiter1997long}: It has been introduced in Section \ref{LSTM}, which is a recurrent neural network with long-term and short-term memory, and has been extensively utilized in time series prediction problems. 
(8) GCNLSTM \cite{cui2019traffic}: It combines graph convolution network with LSTM to capture both spatial correlation (with local neighborhoods) and temporal correlations (short- and long-term time dependency) for prediction.
(9) DKFN \cite{chen2020graph}: It is a novel Kalman Filtering network which can model bias and noise in traffic data, and capture self and neighboring dependencies for prediction. An LSTM is used to model self-dependency and Graph Convolution-LSTM is applied to capture neighbor dependency, and a Kalman Filtering network is utilized to capture the bias and noise in self dependency and neighbor dependency observations rather than treat them as exact ground truth.
(10) MCAN-nhte: It is a ablation study of our method, by removing embedding component in HSC module, (i.e., skip the replacing operation in Figure \ref{correlation_model_details}(b), and directly fed two feature equal length vectors to GCN and LSTM in Figure \ref{correlation_model_details}(a)).
}
Among the baselines, ARIMA, LR, and SVR are well-known for their performance; they are ideal baselines to show the gap between complicated deep learning models and traditional shallow methods. The comparison with them can also reflect the necessity of incorporating spatiotemporal correlations for prediction on a complex real-world dataset. 
{\color{black}
LSTM and GCNLSTM are well-known traditional RNN for time series prediction problem, the comparison with them present the margin of capacity of traditional neural networks and state-of-the-art deep learning methods.
STGCN, BTSP, ASTGCN, and DKFN are state-of-the-art methods whose performances have been evaluated on many real-world datasets. 
These methods relied on different techniques to model the dynamic changes in spatial and/or temporal correlations among road segments for traffic prediction. 
The ablation study will provide insights to the performance and robustness of our method for diverse datasets, i.e., with or without heterogeneous frequency sampling.

}

The comparison between the above methods and our MCAN model demonstrates the effectiveness of different methods for characterizing the dynamic spatiotemporal correlations. 
Note that although there are some works considering multiple types of spatial correlations \cite{geng2019spatiotemporal,pan2019urban}, these
methods are not suitable as baselines because the input (features) suggested in these methods cannot be obtained based on our datasets.

Baselines (1), (2), (3) and (8) don't consider spatial correlations, thus the heterogeneity has little influences to them. 
All the road segments are clustered such that the road segments in the same cluster are with the same observation frequency, and a separate model is built for road segments with the same time frequency.
Baselines (4),(5),(6),(7),(9) (10) and our ablation version MCAN-nhte applied same time frequency (5 minutes) for all road segments in Singapore dataset.
{\color{black}
For Beijing dataset, we test all baselines and our ablation version, using the same time frequency, i.e., 15 minutes. 
It is worth nothing that no weather conditions are included in this dataset, hence we remove both embedding component in HSC module and LSTM in Contextual Factor Module in the ablation version.
}
The hyperparameters of the baselines are optimized using grid search method to ensure a good fit of the data.

\subsubsection{Hyperparameters}

The hyperparameters are tuned using the grid search method. 
The important hyperparameters include the number of the terms of Chebyshev polynomial $K_{em}$ in embedding layers, the $K_{GCN}$ in graph convolution layer (i.e., the size of vector $\mathbf{z}$),
{the size of matrix $M \in \mathbb{R}^{c\times c}$ in embedding component of HSC model}, 
the weight parameters $\alpha$ and $\beta$ in loss function, 
the number of layers in LSTM/FCN modules and the number of neurons in the LSTM/FCN layers, the batch size and the learning rate during the training phase. 
After tuning the hyperparameters through an exhaustive search, the above parameters are set as follows.
Both $K_{em}$ and $K_{GCN}$ are set to 5, the size of matrix $M$ is set to $64\times 64$, the weight parameters in loss function are set to $\alpha = 0.2$ and $\beta = 0.2$ respectively. 
{\color{black}
Each LSTM module contains 3 layers, and each FCN module contains 2 layers. The size of the hidden layers in various LSTM and FCN modules in different components will be presented and discussed in Section \ref{sensitivity_analysis}.}
During the training phase, the batch size is set to 64, and the learning rate is set to 0.0001.
In addition, for incorporating multi-fold temporal correlations, the parameters $lr$, $ld$ and $lw$ are set to 6, 4, and 2, respectively, using grid search based on the prediction performance as well as the available dataset. 
The $k$-fold cross validation is used to test the performance, where $k$ is set to 5. We use leave-one-out cross validation (LOOCV), thus 20\% of the data {\color{black} in test set} is used for validation. The data is shuffled during performance evaluation in order to avoid inherent bias. 
{\color{black}
$H$ refers to the number of time intervals to be predicted. The prediction time horizon of a road segment depends on its size of time interval and $H$, e.g., if $H=6$ and time interval size equals to 5 minutes, the prediction horizon should be $5×6=30$ minutes. For MCAN-nhte and all the other baselines, the time interval applied 5 mins, and for our method MCAN, it depends on the heterogeneous time interval for various road segments. Basically, this paper focuses on short-term prediction, i.e., within a couple of hours (the maximum $H=10$, and the maximum time interval in our dataset is 20 minutes, the max prediction horizon is 200 minutes, less than 4 hours).
}

\subsection{Results}

\subsubsection{Overall Performance.}

\begin{table}[h]
	\centering  
	\fontsize{8pt}{9pt} \selectfont
	\caption{Comparison of results on MAE, MAPE and RMSE on two datasets.}
	\label{PerformanceTable}
	\begin{tabular}{c | c c c | c c c}
		\hline  
		Methods  & \multicolumn{3}{c|}{Our Dataset} & \multicolumn{3}{c}{Q-Traffic Dataset} \\ 
		\cline{2-7}
		 ~ & MAE (km/h) & MAPE (\%) & RMSE (km/h) & MAE (km/h) & MAPE (\%) & RMSE (km/h)\\ \hline 
		ARIMA \cite{williams2003modeling}    & 4.992   & 21.4   & 6.352 & 4.022 & 11.7 & 4.416\\ 
		LR \cite{gross2003linear}       & 4.101   & 19.5   & 4.674 & 1.59 & 7.0 & 2.259 \\ 
		SVR \cite{wu2004travel}      & 3.972   & 18.7   & 4.361 & 5.078 & 21.3 & 5.924 \\
		BTSP \cite{sun2019bus}    & 2.524   & 12.3   & 3.613 & 1.67 & 7.1 & 2.531 \\
		ASTGCN \cite{guo2019attention} & 1.971   & 11.1   & 3.116 & 2.306 & 10.3 & 3.550 \\  
		STGCN \cite{yu2018spatio}  & 3.198   & 13.3   & 4.152 & 2.367 & 10.9 & 3.587 \\ 
		LSTM  \cite{hochreiter1997long} & 1.866 & 12.1 & 2.899 & 1.923 & 8.4 & 2.866 \\
		GCNLSTM \cite{cui2019traffic} &1.714 & 11.1 & 2.719 & 1.831 & 7.9 & 2.715 \\
		DKFN \cite{chen2020graph} & 1.650 & 10.0 & 2.629 & 1.827 & 7.9 & 2.693\\
		\hline %\cline{2-10}
		MCAN-nhte & 1.435 & 9.6 & \textbf{2.021} & \textbf{1.410} & \textbf{6.1} & \textbf{2.116}\\
		MCAN  & \textbf{1.428}  & \textbf{9.4}  & 2.280 & -- & -- & -- \\
		\hline %\cline{2-10}  
	\end{tabular} 
\end{table}

The performance measures used are the Mean Absolute Error (MAE), the Mean Absolute Percentage Error (MAPE), and Root Mean Square Error (RMSE).
Table \ref{PerformanceTable} compares the performance of our proposed method with the baselines {\color{black} on both selected datasets}. 
The traditional time series-based method (ARIMA) obtains poor performance {\color{black} on both datasets}, because it only relies on historical records of traffic speed values without taking into account the impact of spatial correlation and contextual factors. 
{\color{black}
The classic regression methods SVR and LR both marginally outperform ARIMA on our dataset. 
On the Beijing dataset, SVR performs worse than ARIMA, while LR shows good performance.
This indicates that the Beijing dataset in each 15 minutes shows higher linearity than our dataset (due to their data completion technique of dealing with missing data, i.e., using daily average value to fill up), which gives the LR method a better advantage. 
However, they still fail to characterize the complex nonlinear spatiotemporal correlations if the traffic situations change dramatically and dynamically. }
{\color{black}
The LSTM and GCNLSTM learn short-and long-term dependencies that are inherent within speed time series in their architecture. GCLSTM improves the prediction by using GCN to capture spatial correlations. While it outperforms traditional methods, 
GCLSTM still shows lower performance compared with our models (MCAN and MCAN-nhte) on both datasets.
}
The deep learning methods (ASTGCN, STGCN, BTSP) rely on techniques such as graph embedding, GCN to model the dynamic spatial correlations and RNN (LSTM) network to capture the nonlinear temporal correlations; thus they achieve better results than the above-mentioned methods {\color{black} on both datasets}. 
{\color{black}
DKFN shows better performance among deep learning models, as they do not rely on the absolute spatial neighbourhoods and nearby temporal time stamps, but instead introduce noisy measurements to incorporate more realism to real-world datasets, and enables the modelling of data distribution with large fluctuations.
}
Our proposed MCAN outperforms all these deep learn methods as it explicitly models the multi-fold spatial correlations as well as multi-fold temporal correlations, which provide discriminating features to better characterize the dynamic traffic situations. 
In addition, with embedding component, our model settles heterogeneous sampling frequency in data-cube while keeping time consistency by replacing operation and Chebyshev polinomial approximation. 
As such, our method not only sufficiently considers the multi-fold spatiotemperal correlations but efficiently addresses the issues due to the heterogeneous granularity of traffic data. 
{\color{black}
On the other hand, our ablation study with MCAN-nhte also shows top performance on both datasets even without embedding component and LSTM in Contextual Factor Module. This demonstrates that regardless of the datasets and time granularity (homogeneous or heterogeneous sampling frequency, short or long time intervals), our method and even its degraded version, can effectively incorporate multi-fold spatial and temporal correlations to accurately predict traffic situations.}

\begin{figure}[thb]
	\centering
	\includegraphics[width=0.7\textwidth]{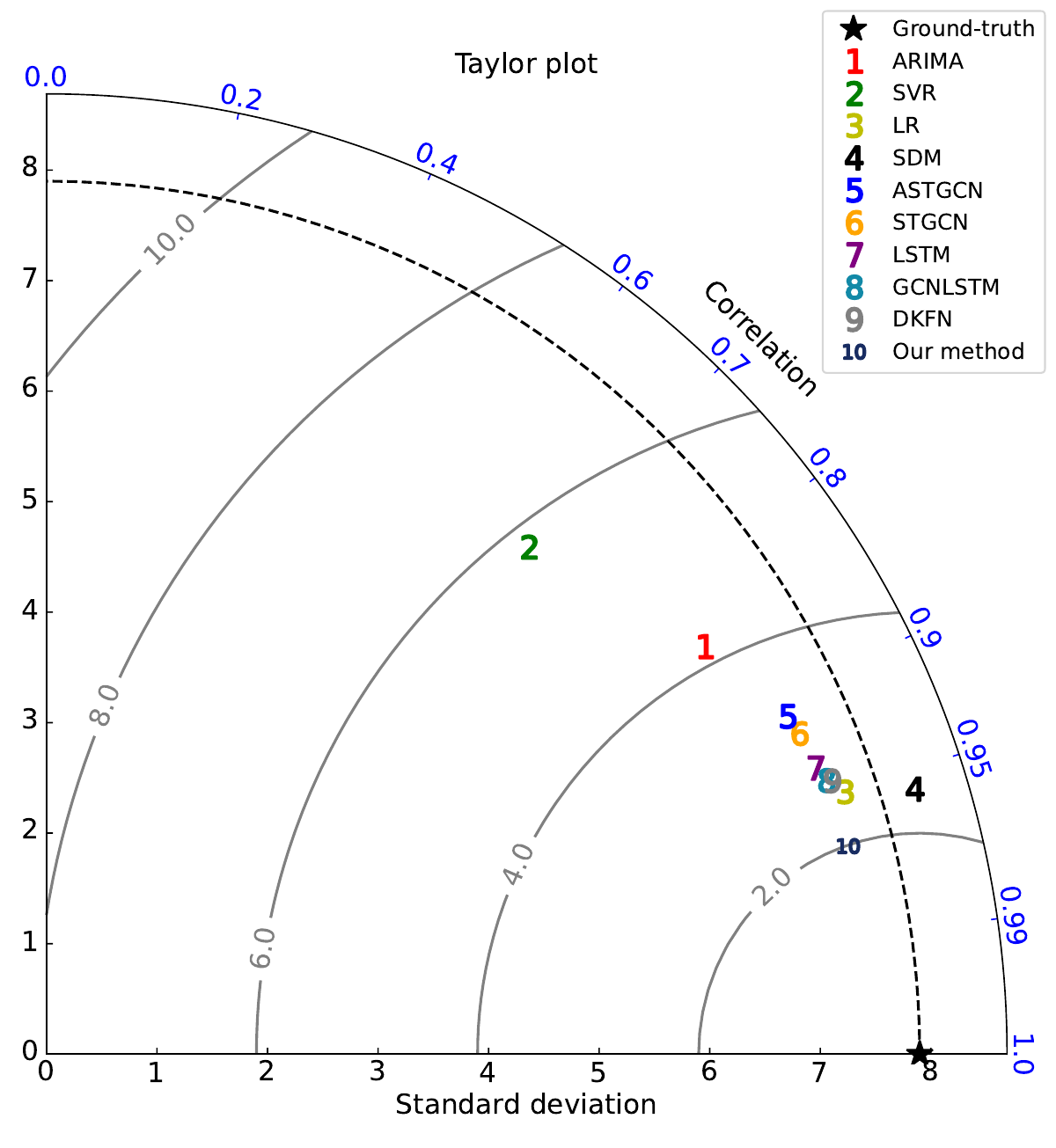} 
	\caption{Taylor plot of our method and baselines.}
	\label{fig_taylor_plot}
\end{figure}
{\color{black}
We also show results in the Taylor plot, which is widely used in climate and other aspects of Earth's environment studies. 
As shown in Figure \ref{fig_taylor_plot}, the black star on the x axis represents the ground truth.
The figure is used to quantify the degree of correspondence between the predicted speeds and ground truths in terms of three statistics: the Pearson correlation coefficient (related to the azimuthal angle), the RMSE (proportional to the distance from the point on the x-axis identified as ``ground truth", i.e., gray contours), and the standard deviation (proportional to the radial distance from the origin, i.e., black dash contours)\footnote{\url{https://en.wikipedia.org/wiki/Taylor_diagram}}. 
The results show that, compared to all the baselines(labeled as `1-9'), our method (labeled as `10') has the smallest RMSE, the highest correlation with the ground truth, and the same standard deviation as the ground truth.
}

\begin{figure}[thb]
	\centering
	\includegraphics[width=0.5\textwidth]{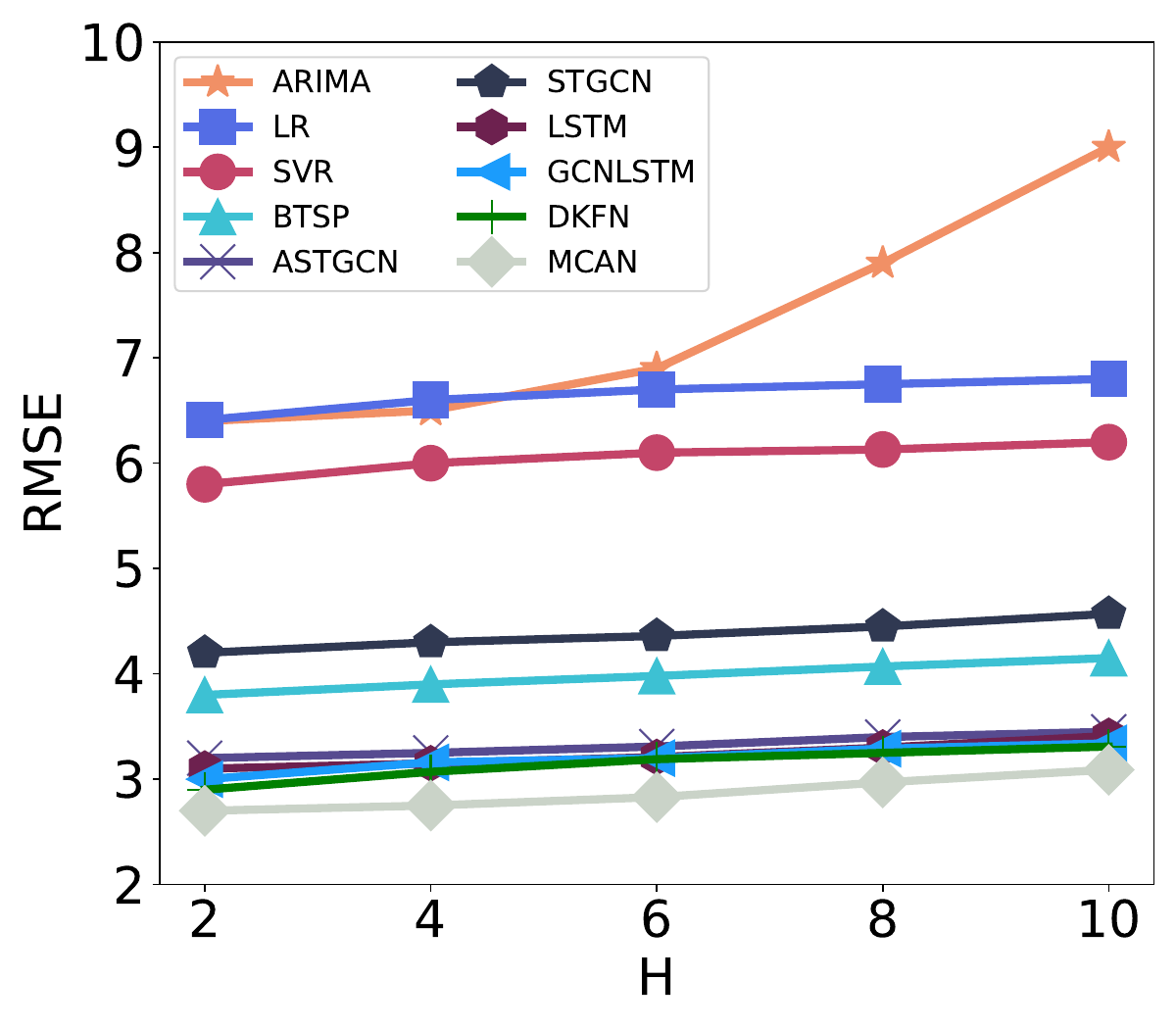} 
	\caption{Changes in performance with the increasing $H$.}
	\label{IncreasingH}
\end{figure}

Figure \ref{IncreasingH} shows the changes of prediction performance of all methods with the increasing $H$ (predict the future $H$ time intervals).
In general, the prediction error becomes larger when $H$ increases for all methods, because the problem becomes more difficult.
The performance of ARIMA degrades dramatically as it only relies on the temporal correlation of recent traffic speeds, thus it can obtain good results for short-term prediction but the performance drops dramatically with the increasing $H$.
The errors of other methods increase slowly with the prediction horizon increases. However, their performance varies significantly, and the deep learning-based methods (BTSP, STGCN, ASTGCN, LSTM, GCNLSTM, DKFN, and MCAN) achieve better results than the traditional methods. This is because they can simultaneously consider the spatial-temporal correlations relying on their respective strategies. 
{\color{black}
The experiments are constructed on Intel(R) Xeon(R) CPU E5-1650 v2 @ 3.50GHz with 32G RAM. 
The training time for our method for 1000 samples, is around 0.4s. The training time for other baselines: ARIMA 1.3s, LR 0.2s, SVR 0.2s, BTSP 1.1s, ASTGCN 0.97s, STGCN 1.5s, LSTM 0.33s, GCNLSTM 0.39s, DKFN 1.2s.}

\begin{figure}
	\centering
	\includegraphics[width=1\textwidth]{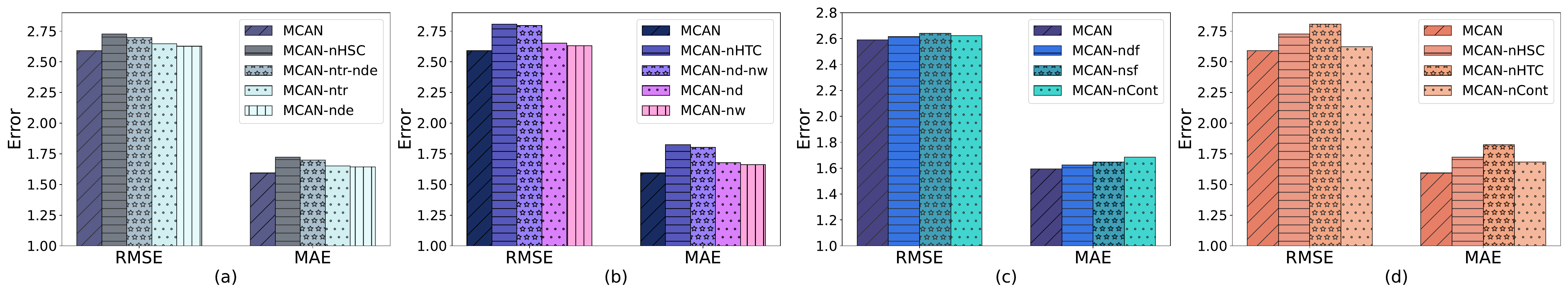}
	\caption{Ablation study (a) effect of multi-fold spatial correlations, (b) effect of multi-fold temporal correlations, (c) evaluation of contextual factors, (d) comparison of different modules. {\color{black} In the experiments, H is set to 1, meaning the prediction horizon depends on the data frequency of each road segment. It can be easily adjusted to a uniform prediction horizon for all roads.}} 
	\label{ParameterEffect}
\end{figure}

\subsubsection{Ablation Study}
In this part, we evaluate the effects of model components that affect the prediction performance, including multi-fold spatial correlations, multi-fold temporal correlations, {\color{black} and the contextual factors module}. 
To verify the effects of \textit{multi-fold spatial correlations}, we designed three degraded versions of MCAN: MCAN-ntr (without spatial correlation of changing trend), MCAN-nde (without spatial correlation of deviation), MCAN-ntr-nde (without spatial correlations of both changing trend and deviation).
Figure \ref{ParameterEffect}(a) shows that removing the multi-fold spatial correlations leads to obvious performance degradation, indicating the necessity to incorporate the multi-fold spatial correlations proposed in this paper. 
{\color{black}
After removing the other two spatial correlations (trend and deviation), and only considering speed observations similar to most of the existing works (MCAN-ntr-nde), the error rates notably increased compared to MCAN. This is because the correlation in terms of speeds sometimes cannot provide essential information, for example, two roads’ speed time series have similar trends but show low linear correlation. In this case, such correlation will not benefit or even has negative impacts on the prediction results. Moreover, adding trend or deviation into consideration is also insufficient (MCAN-ntr and MCAN-nde), because these two correlations alternatively dominate each other. For instance, the MCAN-ntr model will perform unwell when deviation correlation and Pearson correlation are dominated by trend correlation. 
The results also reflect that considering spatial correlations of more folds tends to improve prediction performance. 
It's worth noting that the multiple spatial correlations incorporated could be associated with each other; thus, removing any single spatial correlation does not lead to considerable performance loss.
}
Figure \ref{ParameterEffect}(b) evaluates the impact of \textit{multi-fold temporal correlations} by comparing with the following degraded version:  MCAN-nd (without temporal correlation of daily periodicity), MCAN-nw (without temporal correlation of weekly periodicity), MCAN-nd-nw (without temporal correlations of both daily and weekly periodicity). The results reflect that the consideration of multi-fold temporal correlations is important. 
{\color{black}
The explanation is as follows. Sometimes recent observations fail to depict inherent traffic patterns. For instance, even though the traffic states follow the daily periodicity  congestion in the upcoming peak hours, the recent observations have strong fluctuation and high noise due to data collection problems. As such, the recent observations will not provide informative features for the model. Considering daily periodicity will provide an essential clue for a good inference. Moreover, Figure \ref{ParameterEffect}(b) reveals that only considering daily or weekly periodicity cannot sufficiently complement the recent dependency for good prediction: MCAN-nd and MCAN-nw slightly outperform the MCAN-nd-nwh version, with a higher error rate than MCAN. These results validate the effect of multi-fold temporal correlation module. 

To verify the effects of \textit{contextual factors module}, we designed three degraded versions of MCAN: MCAN-ndf (without dynamic factors such as weather conditions), MCAN-nsf (without static factors), MCAN-ncont (without contextual factors of both dynamic and static factors). 
As shown in Figure \ref{ParameterEffect}(c), the results reflect that the absence of contextual factors increased the prediction error rate moderately, which indicates that the weather, calendar, and road characteristics slightly affect the prediction results. 
This is because the weather conditions in a moderately large region are usually the same. Except for rare extreme weather, the changes also vary slowly. The road characteristics basically do not change over time, therefore, their impacts are not very important. 
Figure \ref{ParameterEffect}(d) compares the influences of different modules of MCAN, by removing HTC module, HSC module, and contextual factor module, separately. In Figure 9 (d) we present three ablation versions, MCAN-nHSC, MCAN-nHTC, and MCAN-nCont. The results show the obvious notable increase in prediction errors when removing HTC and HSC components, while a mild increase when removing contextual factor module.
}

\subsubsection{Sensitivity Analysis.}
\label{sensitivity_analysis}

\begin{figure}[thb]
	\centering
	\includegraphics[width=1\textwidth]{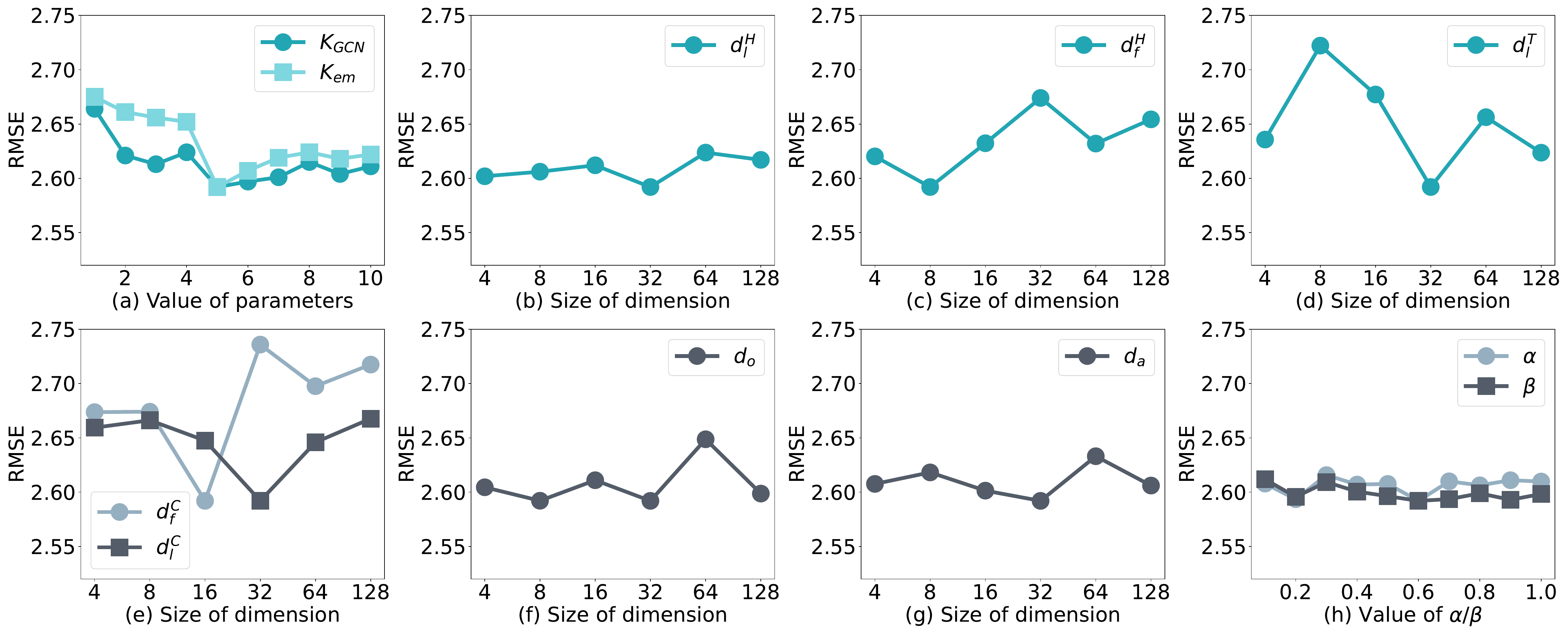} 
	\caption{Sensitivity analysis.}
	\label{sensi_analysis}
\end{figure}

{\color{black}
Figure.\ref{sensi_analysis} (a) illustrates the effects of the number of $K_{GCN}$ and $K_{em}$. The results show that the RMSE first decreases with the increasing $K_{GCN}$, then it tends to increase if $K_{GCN}$ continues to increase. This is because, more items of CPA can better learn the complex traffic situations of local neighbors, leading to a higher ability of approximating traffic speed distribution of different road segments with diverse time granularity. As such, increasing $K_{GCN}$ improves the prediction performance when $K_{GCN}$ is relatively small. However, the prediction performance tends to degrade if the $K_{GCN}$ continues to increase when $K_{GCN}$ exceeds 5. The main reason is that the size of the model grows rapidly with redundant parameters, and the model becomes overfitting during training for the dataset. Similar trends are observed for parameter $K_{em}$.

Next, we will introduce other hyperparameters in Figure \ref{sensi_analysis} (b)-(g). 
In Multi-fold Spatial Correlation module, all the LSTM components are 3-layers LSTM with the same hidden size $d_l^H$, and all the FCN are 2-layers with the same hidden size $d_f^H$. 
In Multi-fold Temporal Correlation module, all the LSTM are 3-layers with the same hidden size $d_l^T$. In Contextual module, all the LSTMs are 3-layers with the same hidden size $d_l^C$ and all the FCN are 2-layers with the same hidden size $d_f^C$. 
The intermediate output feature vectors $h^{tr}$, $h^{de}$, $h^{S}$, $h^{recent}$, $h^{daily}$, $h^{weekly}$, $h^{E}$, and $h^{R}$ are of the same length $d_o$.
We also test the attention layer, which is a two-layers FCN with hidden size $d_a$. 
For the prediction layer, we simply employ a one-layer FCN with the output size equal to 1. 
Based on \cite{pan2019urban}, we conduct grid search on the number of all above hidden/output size over {4, 8, 16, 32, 64, 128}. 
As shown in Figure \ref{sensi_analysis} (b), (d), and (e), $d_l^H$, $d_l^T$, and $d_l^C$, the hidden sizes of the LSTM in three various modules, show the similar trend: i.e., all achieve the best results when they equal to 32, and have higher error rate when the size is too small or too large. This means that in our problem, when learning features with temporal dependencies, an LSTM with a few parameters cannot learn such complex dynamics, while an LSTM with too many parameters will not be trained effectively with limited samples. 
In Figure \ref{sensi_analysis} (c) and (e), $d_f^H$, $d_f^C$ equals to 8, 16, respectively, for the lowest RMSE, and with the increasing dimension size, the error rate increased as well. The effect of FCN in MSC module (determined by $d_f^H$) is to project multi-fold spatial correlations into the same latent space with the other modules, the FCN in Contextual module aims to learn from static features and then project them into latent space as well (determined by $d_f^C$). 
Both FCNs can achieve good results just with low parameter number (8 and 16 hidden size), while too many neurons will lead to unnecessary redundancy for model learning capacity.
Figure \ref{sensi_analysis} (g) and (f) illustrate the effect of $d_a$ and $d_o$. Compared to (c), (d) and (e), their error rates changed in a narrow range, which indicates the their importance is not as high as the others. This is because most of the essential information which play an important role in prediction have been successfully incorporated into our intermediate outputs (i.e., $h^{tr}$, $h^{de}$, $h^{S}$, $h^{recent}$, $h^{daily}$, $h^{weekly}$, $h^{E}$, and $h^{R}$). 

Figure \ref{sensi_analysis}(h) evaluates the effects of $\alpha$ and $\beta$, the weights of last two terms in the loss function as equation \ref{Loss}. 
$\alpha$ determines the weight of the loss of changing trend, i.e., the accuracy of HSC-trend component. 
Similarly, $\beta$ determines the weight of loss of deviation. 
Figure \ref{sensi_analysis}(h) shows that the error rates of both parameters fluctuate in a relatively narrow range ($[2.550,2.600]$ in RMSE) compared to the other parameters. This indicates that the most important term in loss function is still the $\parallel \hat{y_i}-y_i\parallel^2$. Also, $\alpha$ and $\beta$ achieve the lowest RMSE when they are equal to 0.2. 
The large weights will cause the loss function to pay unnecessary attention to the last two terms rather than the first one.

}

{\color{black}
\subsubsection{Case Study.}
\begin{figure}[thb]
	\centering
	\includegraphics[width=0.7\textwidth]{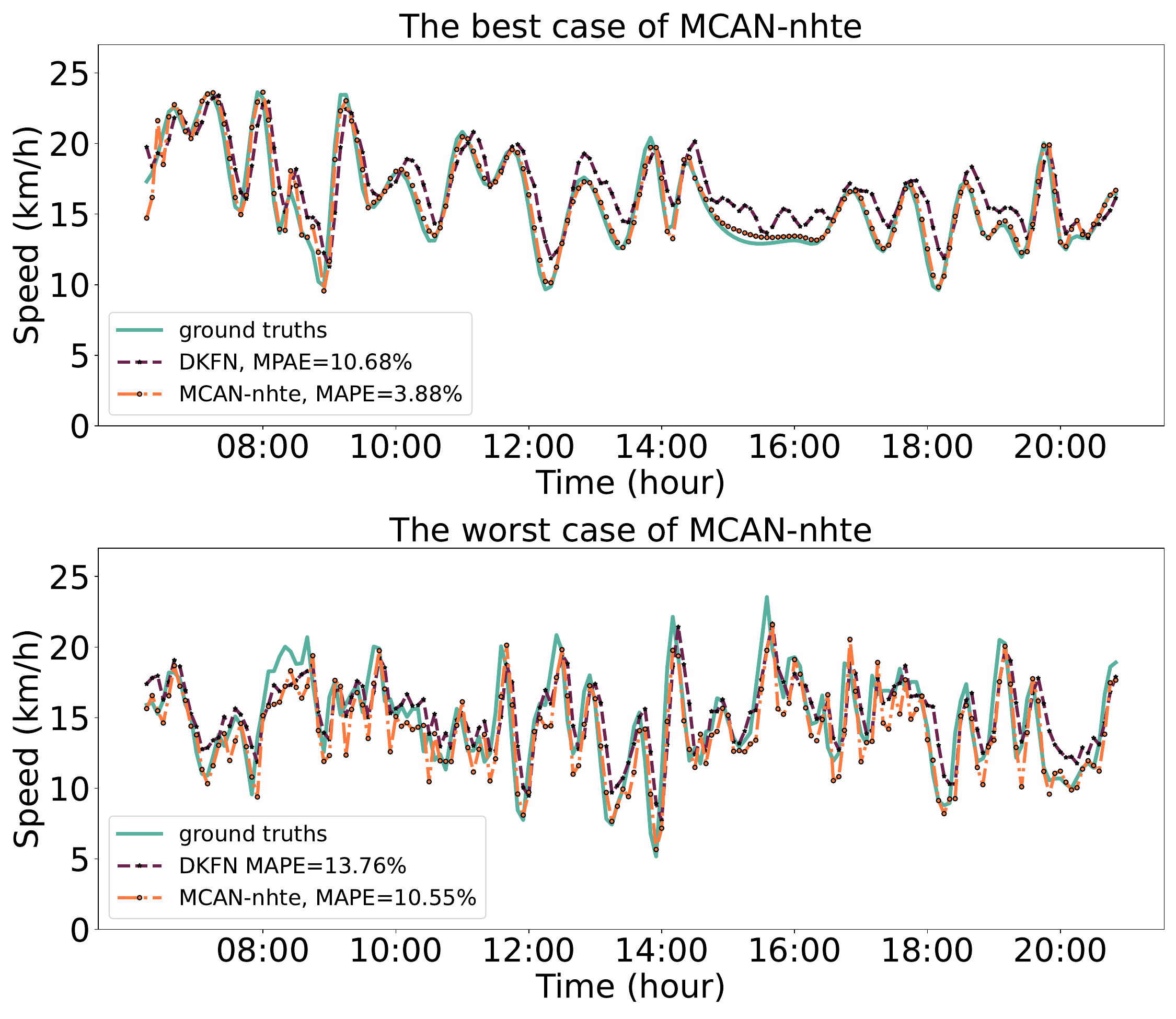} 
	\caption{In our dataset in Singapore, the best and the worst cases study of our method and the second-best baseline “DKFN”. H=1, the time interval is 5 minutes.}
	\label{casestudy_1}
\end{figure}

In Figure \ref{casestudy_1} to Figure \ref{casestudy_4}, we studied the best and worst cases of our method, compared with the second-best baseline on two datasets, under various traffic situations. 
As shown in Figure \ref{casestudy_1}, the green solid line is the ground-truth traffic speed, the orange dash-dot line is our prediction results using MCAN-nhte, and the purple dash line is the prediction of DKFN. The prediction horizon H=1, time interval size is 5 minutes, and both methods applied the same time interval. Based on observation, in the best case, our MCAN-nhte has $MAPE=3.88\%$, while DKFN has $MAPE=10.68\%$. Our method fit the ground-truth very well, while DKFN underperforms during fluctuations (e.g., 10:00-12:00, 15:00-16:00, 18:00-20:00). In the worst case, our MCAN-nhte with $MAPE=10.55\%$ is moderately higher than DKFN. We can observe that while the ground truth speed distribution in the worst case exhibits more fluctuations than in the best case, our method still fits the ground-truth well even during extreme changes in traffic situations (e.g. around 10:00, 12:00, 14:00, and 18:00).

\begin{figure}[thb]
	\centering
	\includegraphics[width=0.7\textwidth]{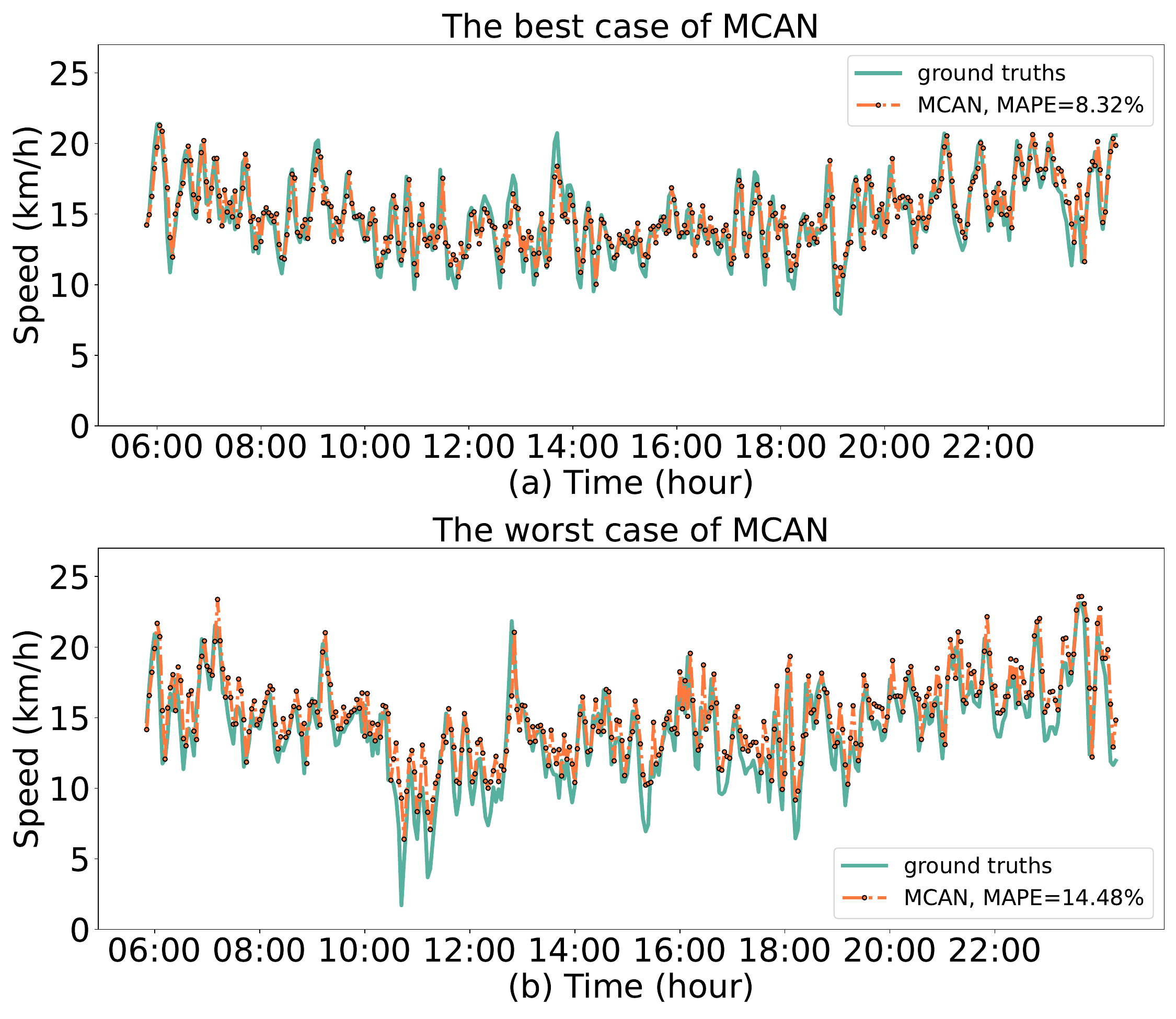} 
	\caption{In our dataset in Singapore, the best and the worst cases study of our method applied heterogeneous time granularity, time interval=3min, H=1.}
	\label{casestudy_2}
\end{figure}
Figure \ref{casestudy_2} presents our best and worst cases of MCAN in fine time granularity. We do not present the second-best baseline results here because all baselines are designed for homogeneous time granularity. However, based on the time interval of 3 minutes, it can be observed that the traffic speeds fluctuate highly and frequently, which indicates the effectiveness and necessity of our method MCAN (considering heterogeneous time granularity). Some road segments are located in important and busy areas, and predicting average traffic speed in the next large time interval (e.g. 15 minutes) will not provide useful information to the users. For example, a driver may want to know the traffic situation in the turning ahead for route planning (i.e., to proceed with the turn or detour from potential congestion). Our method can provide the prediction result within the next minute rather than just providing an average speed for the next 15 minutes. The results show that in the best case, our method can predict traffic speed accurately (with $MAPE=8.32\%$). In the worst case, it can also capture the main changing trends in speed time series very well (with $MAPE=14.48\%$). 

\begin{figure}[thb]
	\centering
	\includegraphics[width=0.7\textwidth]{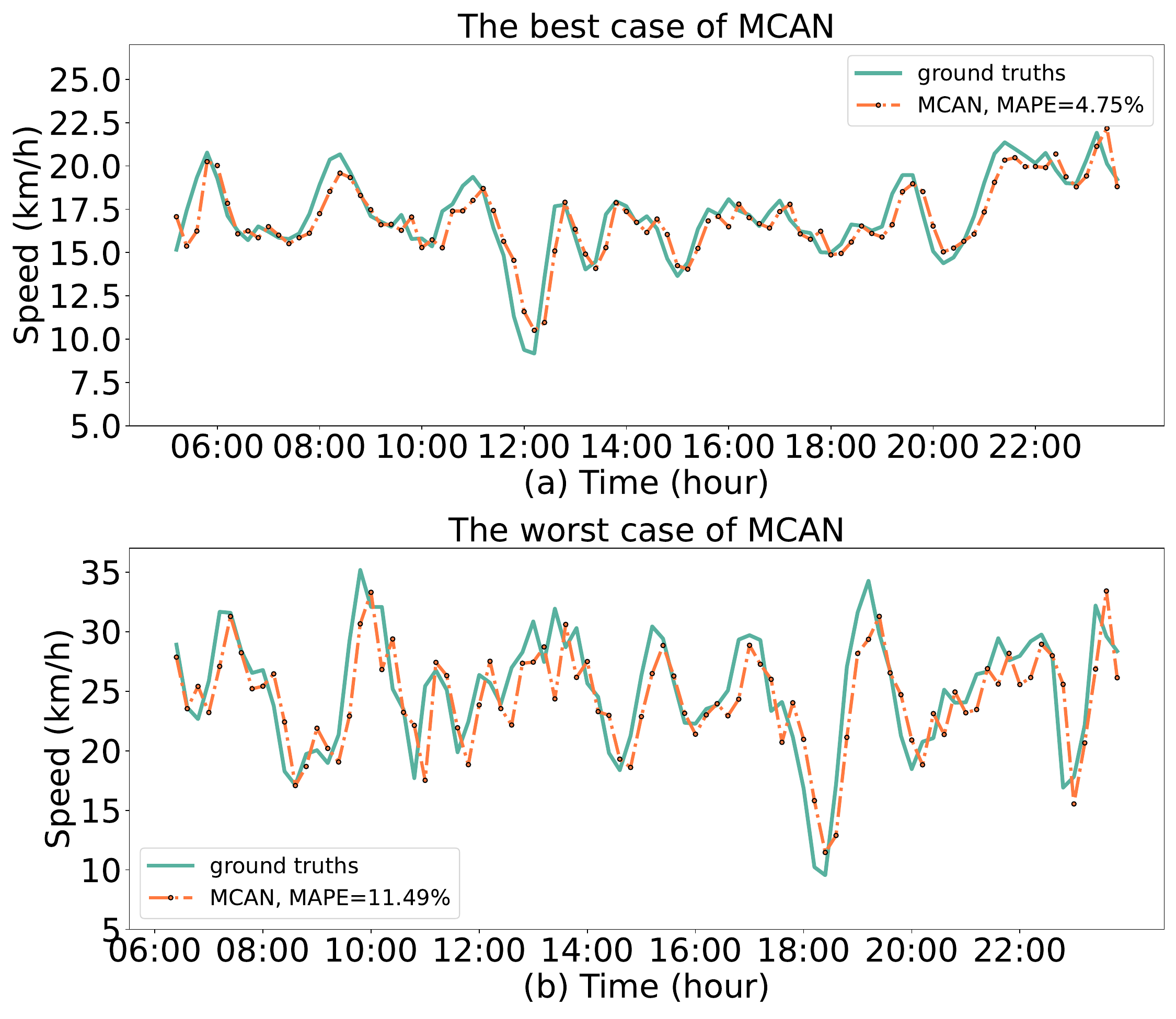} 
	\caption{In our dataset in Singapore, the best and the worst cases study of our method applied heterogeneous time granularity, time interval=12min, H=1.}
	\label{casestudy_3}
\end{figure}

In addition, for road segments located in the non-congested area, the traffic situations can change slowly and smoothly. Therefore, setting the time interval to 10-15 minutes is sufficient. For these roads, extraction of coarse time granularity is effective and low cost. Figure 14 demonstrates that our method is also accurate for a large time interval speed time series. It can achieve $MAPE=4.75\%$ in the best case and $11.49\%$ in the worst case. 

\begin{figure}[thb]
	\centering
	\includegraphics[width=0.7\textwidth]{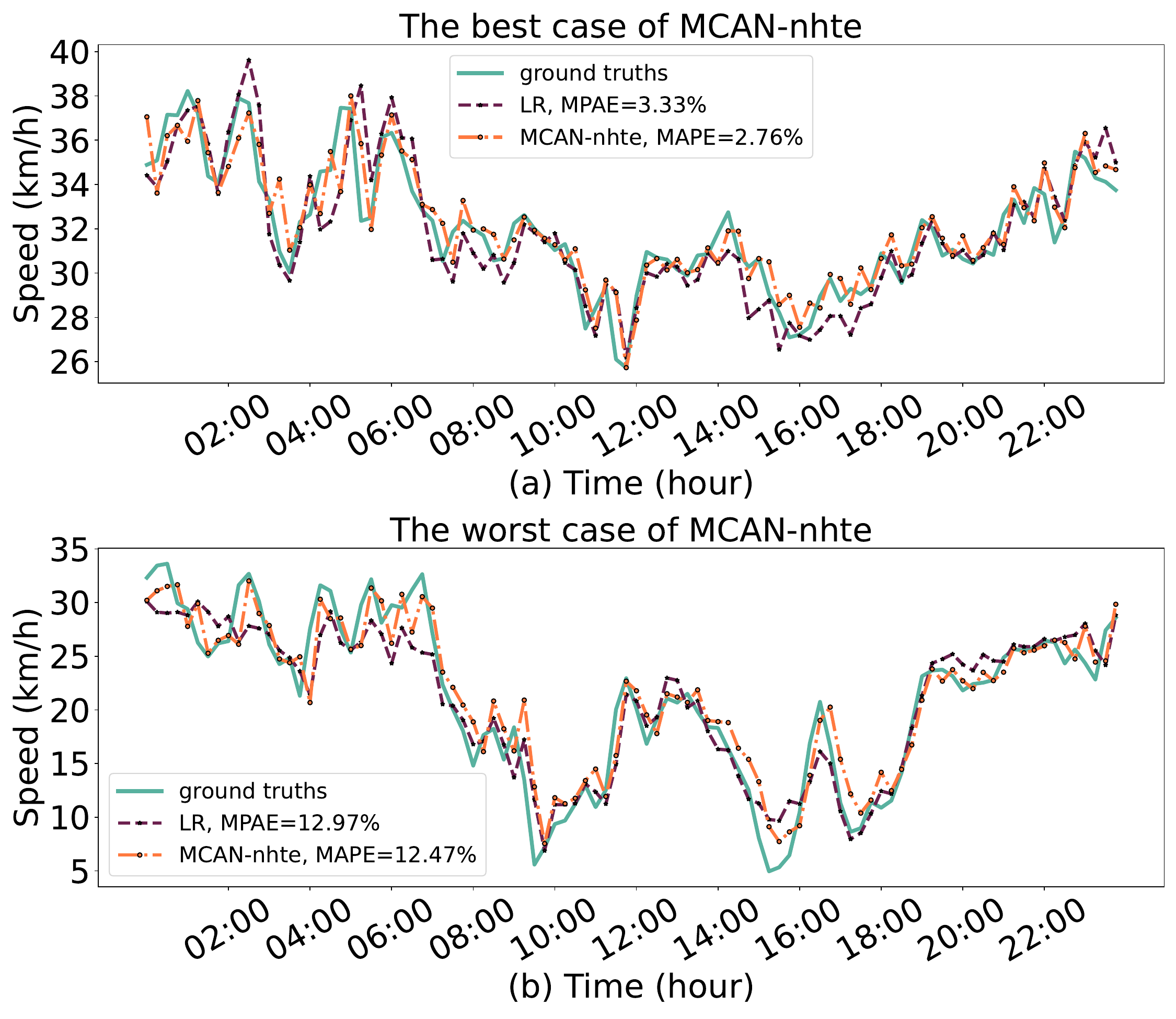} 
	\caption{Figure. 15. In Q Traiffic dataset in Beijing, the best and the worst cases study of our method and the second-best baseline LR. The time interval=15min, H=1.}
	\label{casestudy_4}
\end{figure}
 In Figure \ref{casestudy_4}, the green solid line is the ground-truth traffic speed, the orange dash-dot line is our prediction using MCAN-nhte, and the purple dash line is the prediction of second-best baseline LR. In the other dataset, our method with degraded version (MCAN-nhte) also shows very good performance compared to the second-best baseline (LR) for both the best case and the worst case. This demonstrates that our method is feasible and accurate for the various datasets considered.
}

\section{Conclusion}
{\color{black} 
This paper addresses the short-term traffic prediction problem, which is challenging due to severe fluctuations in traffic data. We demonstrate that traffic data exhibit multi-fold correlations which provide new opportunities to improve traffic prediction accuracy. We developed a multi-fold correlation attention network called MCAN to achieve accurate prediction, by learning and incorporating the multi-fold spatial correlations, multi-fold temporal correlations, and contextual factors.
We also showed that utilizing heterogeneous frequency traffic data, instead of transforming them into coarse-grained data of uniform frequency, leads to significant performance gains. We have evaluated our methods on two real-world datasets and undertook extensive comparisons with several state-of-the-art methods.

The proposed MCAN is capable of learning multi-fold spatiotemporal correlations and is suitable for tackling challenging graph-structured data prediction problems, e.g., data exhibiting severe fluctuations.
Since the proposed MCAN is an end-to-end deep neural network model, it suffers from the common drawbacks of other deep learning models. 
For example, the hyper-parameters tuning process depends mainly on the characteristics of the target dataset; thus, additional adaptations are required to achieve satisfactory performance on different datasets. 
In particular, to tackle the challenges of short-term traffic prediction, MCAN incorporates multiple spatial and temporal correlation modules to provide discriminating features to better characterize dynamic traffic situations.
Nevertheless, this also significantly increases the model complexity, which requires larger training data and efforts for model tuning.  
}

{\color{black}
In this paper, the grid search has been used to optimize the hyperparameters of MCAN. In the future, we plan to further improve our MCAN by combining it with optimization algorithms such as gravitation search algorithm (GSA) or genetic algorithm (GA). In addition, neural architecture search (NAS) \cite{elsken2019neural} has been extensively studied for automating the design of deep neural networks by searching the optimal network architecture. However, NAS methods typically assume fixed hyperparameters while hyperparameter optimization assumes fixed network architecture, i.e., the two problems are solved separately. We plan to jointly optimize neural architectures and hyperparameters as part of our future work.
{\color{black}
In addition, many works proposed novel GCN structures to capture dynamic spatial correlations intelligently, which could be employed to further improve our model. For example, optimizing the fixed Laplace matrix as \cite{guo2020optimized}, or enabling the method to be diffusion-aware for better capturing traffic evolution  \cite{chen2019gated,zhang2021traffic,chen2022applying}. 
However, the existing GCN techniques are typically time-consuming (for capturing complex and dynamic correlations); thus, incorporating such a design will significantly increase the model complexity. Thus, novel GCN techniques should be developed to balance efficiency and performance. 
}

}

{\color{black} 
Moreover, we also plan to apply the multi-fold correlation attention network to predict traffic speeds on large-scale traffic networks. To achieve this, several challenges need to be addressed. 
For example, with a large number of road segments, the trade-off between accuracy and efficiency should be considered for platforms with limited computing resources. 
Also, it is usually difficult to obtain high-quality traffic data over the entire traffic network. 
Thus, techniques need to be developed for handling low-quality data such as heterogeneous type/structure, sparsity, etc.
}

\section{Acknowledgments}
This research project is supported in part by the National Research Foundation Singapore under its Campus for Research Excellence and Technological Enterprise (CREATE) programme with the Technical University of Munich at TUMCREATE.

% \section*{References}
\bibliographystyle{plain}
\bibliography{mybibfile}

\begin{thebibliography}{10}

\bibitem{abdelraouf2021utilizing}
Amr Abdelraouf, Mohamed Abdel-Aty, and Jinghui Yuan.
\newblock Utilizing attention-based multi-encoder-decoder neural networks for
  freeway traffic speed prediction.
\newblock {\em IEEE Trans. Intell. Transp. Syst.}, 2021.

\bibitem{alsolami2020hybrid}
Bdoor Alsolami, Rashid Mehmood, and Aiiad Albeshri.
\newblock Hybrid statistical and machine learning methods for road traffic
  prediction: a review and tutorial.
\newblock In {\em Smart Infrastr. Appl.}, pages 115--133. Springer, 2020.

\bibitem{asadi2020spatio}
Reza Asadi and Amelia~C Regan.
\newblock A spatio-temporal decomposition based deep neural network for time
  series forecasting.
\newblock {\em Appl. Soft Comput.}, 87:105963, 2020.

\bibitem{bokde2018analysis}
Neeraj Bokde, Andr{\'e}s Feij{\'o}o, and Kishore Kulat.
\newblock Analysis of differencing and decomposition preprocessing methods for
  wind speed prediction.
\newblock {\em Appl. Soft Comput.}, 71:926--938, 2018.

\bibitem{chai2018bike}
Di~Chai, Leye Wang, and Qiang Yang.
\newblock Bike flow prediction with multi-graph convolutional networks.
\newblock In {\em 26th ACM SIGSPATIAL Int. Conf. Adv. Geogr. Inf. Syst.
  (SIGSPATIAL)}, pages 397--400. ACM, 2018.

\bibitem{chang2018structure}
Jianlong Chang, Jie Gu, Lingfeng Wang, Gaofeng Meng, Shiming Xiang, and
  Chunhong Pan.
\newblock Structure-aware convolutional neural networks.
\newblock In {\em Adv. Neural Inf. Process. Syst. (NeurlPS)}, pages 11--20,
  2018.

\bibitem{chen2019gated}
Cen Chen, Kenli Li, Sin~G Teo, Xiaofeng Zou, Kang Wang, Jie Wang, and Zeng
  Zeng.
\newblock Gated residual recurrent graph neural networks for traffic
  prediction.
\newblock In {\em Thirty-Third AAAI Conf. Artif. Intell. (AAAI)}, volume~33,
  pages 485--492, 2019.

\bibitem{chen2020graph}
Fanglan Chen, Zhiqian Chen, Subhodip Biswas, Shuo Lei, Naren Ramakrishnan, and
  Chang-Tien Lu.
\newblock Graph convolutional networks with kalman filtering for traffic
  prediction.
\newblock In {\em 28th Int. Conf. Adv. Geographic Inf. Syst.}, pages 135--138,
  2020.

\bibitem{chen2022applying}
Gen Chen and Jiawan Zhang.
\newblock Applying artificial intelligence and deep belief network to predict
  traffic congestion evacuation performance in smart cities.
\newblock {\em Appl. Soft Comput.}, page 108692, 2022.

\bibitem{cui2019traffic}
Zhiyong Cui, Kristian Henrickson, Ruimin Ke, and Yinhai Wang.
\newblock Traffic graph convolutional recurrent neural network: A deep learning
  framework for network-scale traffic learning and forecasting.
\newblock {\em IEEE Trans. Intell. Transp. Syst.}, 21(11):4883--4894, 2019.

\bibitem{cui2016deep}
Zhiyong Cui, Ruimin Ke, Yinhai Wang, et~al.
\newblock Deep stacked bidirectional and unidirectional lstm recurrent neural
  network for network-wide traffic speed prediction.
\newblock In {\em 6th Int. Workshop Urban Comput. (UrbComp)}, 2016.

\bibitem{dai2016discriminative}
Hanjun Dai, Bo~Dai, and Le~Song.
\newblock Discriminative embeddings of latent variable models for structured
  data.
\newblock In {\em Thirty-Third Int. Conf. Mach. Learn. (ICML)}, pages
  2702--2711, 2016.

\bibitem{deng2016latent}
Dingxiong Deng, Cyrus Shahabi, Ugur Demiryurek, Linhong Zhu, Rose Yu, and Yan
  Liu.
\newblock Latent space model for road networks to predict time-varying traffic.
\newblock In {\em 22th ACM SIGKDD Int. Conf. Knowl. Discov. Data Min.}, pages
  1525--1534. ACM, 2016.

\bibitem{du2019deep}
Bowen Du, Hao Peng, Senzhang Wang, Md~Zakirul~Alam Bhuiyan, Lihong Wang, Qiran
  Gong, Lin Liu, and Jing Li.
\newblock Deep irregular convolutional residual lstm for urban traffic
  passenger flows prediction.
\newblock {\em IEEE Trans. Intell. Transp. Syst.}, 21(3):972--985, 2019.

\bibitem{duan2016starima}
Peibo Duan, Guoqiang Mao, Changsheng Zhang, and Shangbo Wang.
\newblock Starima-based traffic prediction with time-varying lags.
\newblock In {\em 16th IEEE Int. Conf. Intell. Transp. Syst. (ITSC)}, pages
  1610--1615. IEEE, 2016.

\bibitem{elsken2019neural}
Thomas Elsken, Jan~Hendrik Metzen, and Frank Hutter.
\newblock Neural architecture search: A survey.
\newblock {\em J. Mach. Learn. Res.}, 20(1):1997--2017, 2019.

\bibitem{fang2019gstnet}
Shen Fang, Qi~Zhang, Gaofeng Meng, Shiming Xiang, and Pan Chunhong.
\newblock Gstnet: Global spatial-temporal network for traffic flow prediction.
\newblock In {\em Thirty-Third AAAI Conf. Artif. Intell. (AAAI)}, pages
  2286--2293, 2019.

\bibitem{feng2018adaptive}
Xinxin Feng, Xianyao Ling, Haifeng Zheng, Zhonghui Chen, and Yiwen Xu.
\newblock Adaptive multi-kernel svm with spatial--temporal correlation for
  short-term traffic flow prediction.
\newblock {\em IEEE Trans. Intell. Transp. Syst.}, 20(6):2001--2013, 2018.

\bibitem{geng2019spatiotemporal}
Xu~Geng, Yaguang Li, Leye Wang, Lingyu Zhang, Qiang Yang, Jieping Ye, and Yan
  Liu.
\newblock Spatiotemporal multi-graph convolution network for ride-hailing
  demand forecasting.
\newblock In {\em Thirty-Third AAAI Conf. Artif. Intell. (AAAI)}, volume~33,
  pages 3656--3663, 2019.

\bibitem{gross2003linear}
Jurgen Gross and J{\"u}rgen Gro{\ss}.
\newblock {\em Linear regression}, volume 175.
\newblock Springer Science \& Business Media, 2003.

\bibitem{guo2020optimized}
Kan Guo, Yongli Hu, Zhen Qian, Hao Liu, Ke~Zhang, Yanfeng Sun, Junbin Gao, and
  Baocai Yin.
\newblock Optimized graph convolution recurrent neural network for traffic
  prediction.
\newblock {\em IEEE Trans. Intell. Trans. Syst.}, 22(2):1138--1149, 2020.

\bibitem{guo2021hierarchical}
Kan Guo, Yongli Hu, Yanfeng Sun, Sean Qian, Junbin Gao, and Baocai Yin.
\newblock Hierarchical graph convolution networks for traffic forecasting.
\newblock In {\em AAAI Conf. Artif. Intell.}, volume~35, pages 151--159, 2021.

\bibitem{guo2019attention}
Shengnan Guo, Youfang Lin, Ning Feng, Chao Song, and Huaiyu Wan.
\newblock Attention based spatial-temporal graph convolutional networks for
  traffic flow forecasting.
\newblock In {\em Thirty-Third AAAI Conf. Artif. Intell. (AAAI)}, volume~33,
  pages 922--929, 2019.

\bibitem{han2021dynamic}
Liangzhe Han, Bowen Du, Leilei Sun, Yanjie Fu, Yisheng Lv, and Hui Xiong.
\newblock Dynamic and multi-faceted spatio-temporal deep learning for traffic
  speed forecasting.
\newblock In {\em 27th ACM SIGKDD Int. Conf. Knowl. Discov. Data Min.}, pages
  547--555, 2021.

\bibitem{henaff2015deep}
Mikael Henaff, Joan Bruna, and Yann LeCun.
\newblock Deep convolutional networks on graph-structured data.
\newblock {\em arXiv preprint arXiv:1506.05163}, 2015.

\bibitem{hochreiter1997long}
Sepp Hochreiter and J{\"u}rgen Schmidhuber.
\newblock Long short-term memory.
\newblock {\em Neural Comput.}, 9(8):1735--1780, 1997.

\bibitem{hou2016repeatability}
Zhongsheng Hou and Xingyi Li.
\newblock Repeatability and similarity of freeway traffic flow and long-term
  prediction under big data.
\newblock {\em IEEE Trans. Intell. Transp. Syst.}, 17(6):1786--1796, 2016.

\bibitem{hu2016crowdsourcing}
Huiqi Hu, Guoliang Li, Zhifeng Bao, Yan Cui, and Jianhua Feng.
\newblock Crowdsourcing-based real-time urban traffic speed estimation: From
  trends to speeds.
\newblock In {\em 2016 IEEE 32nd Int. Conf. Data Eng. (ICDE)}, pages 883--894.
  IEEE, 2016.

\bibitem{khosravi2011genetic}
Abbas Khosravi, Ehsan Mazloumi, Saeid Nahavandi, Doug Creighton, and JWC
  Van~Lint.
\newblock A genetic algorithm-based method for improving quality of travel time
  prediction intervals.
\newblock {\em Transport. Res. C-Emerg. Technol.}, 19(6):1364--1376, 2011.

\bibitem{koller2009probabilistic}
Daphne Koller and Nir Friedman.
\newblock {\em Probabilistic graphical models: principles and techniques}.
\newblock MIT press, 2009.

\bibitem{liao2018deep}
Binbing Liao, Jingqing Zhang, Chao Wu, Douglas McIlwraith, Tong Chen, Shengwen
  Yang, Yike Guo, and Fei Wu.
\newblock Deep sequence learning with auxiliary information for traffic
  prediction.
\newblock In {\em 24th ACM SIGKDD Int. Conf. Knowl. Discov. Data Min.}, pages
  537--546, 2018.

\bibitem{luo2019spatiotemporal}
Xianglong Luo, Danyang Li, Yu~Yang, and Shengrui Zhang.
\newblock Spatiotemporal traffic flow prediction with knn and lstm.
\newblock {\em J. Adv. Transport.}, 2019, 2019.

\bibitem{mckinley1998cubic}
Sky McKinley and Megan Levine.
\newblock Cubic spline interpolation.
\newblock {\em Coll. Redwoods}, 45(1):1049--1060, 1998.

\bibitem{mir2016adaptive}
Zeeshan~Hameed Mir and Fethi Filali.
\newblock An adaptive kalman filter based traffic prediction algorithm for
  urban road network.
\newblock In {\em 2016 12th Int. Conf. Innova. Inf. Technol. (IIT)}, pages
  1--6. IEEE, 2016.

\bibitem{oh2015improvement}
Simon Oh, Young-Ji Byon, and Hwasoo Yeo.
\newblock Improvement of search strategy with k-nearest neighbors approach for
  traffic state prediction.
\newblock {\em IEEE Trans. Intell. Transp. Syst.}, 17(4):1146--1156, 2015.

\bibitem{pan2019urban}
Zheyi Pan, Yuxuan Liang, Weifeng Wang, Yong Yu, Yu~Zheng, and Junbo Zhang.
\newblock Urban traffic prediction from spatio-temporal data using deep meta
  learning.
\newblock In {\em 25th ACM SIGKDD Int. Conf. Knowl. Discov. Data Min.
  (SIGKDD)}, pages 1720--1730, 2019.

\bibitem{peng2019frequency}
Shunfeng Peng, Yanyan Shen, Yanmin Zhu, and Yuting Chen.
\newblock A frequency-aware spatio-temporal network for traffic flow
  prediction.
\newblock In {\em Int. Conf. Database Syst. Adv. Appl. (DASFFAA)}, pages
  697--712. Springer, 2019.

\bibitem{schulz2018tutorial}
Eric Schulz, Maarten Speekenbrink, and Andreas Krause.
\newblock A tutorial on gaussian process regression: Modelling, exploring, and
  exploiting functions.
\newblock {\em J. Math. Psychol.}, 85:1--16, 2018.

\bibitem{sun2019bus}
Yidan Sun, Guiyuan Jiang, Siew-Kei Lam, Shicheng Chen, and Peilan He.
\newblock Bus travel speed prediction using attention network of heterogeneous
  correlation features.
\newblock In {\em 2019 SIAM Int. Conf. Data Min. (SDM)}, pages 73--81. SIAM,
  2019.

\bibitem{van2012short}
JWC Van~Lint and CPIJ Van~Hinsbergen.
\newblock Short-term traffic and travel time prediction models.
\newblock {\em Artif. Intell. Appl. Crit. Transp. Iss.}, 22(1):22--41, 2012.

\bibitem{wang2016traffic}
Jingyuan Wang, Qian Gu, Junjie Wu, Guannan Liu, and Zhang Xiong.
\newblock Traffic speed prediction and congestion source exploration: A deep
  learning method.
\newblock In {\em 2016 IEEE Int. Conf. Data Min. (ICDM)}, pages 499--508. IEEE,
  2016.

\bibitem{wang2014real}
Miao Wang, Hangguan Shan, Rongxing Lu, Ran Zhang, Xuemin Shen, and Fan Bai.
\newblock Real-time path planning based on hybrid-vanet-enhanced transportation
  system.
\newblock {\em IEEE Trans. Veh. Technol.}, 64(5):1664--1678, 2014.

\bibitem{williams2003modeling}
Billy~M Williams and Lester~A Hoel.
\newblock Modeling and forecasting vehicular traffic flow as a seasonal arima
  process: Theoretical basis and empirical results.
\newblock {\em J. Transp. Eng.}, 129(6):664--672, 2003.

\bibitem{wu2004travel}
Chun-Hsin Wu, Jan-Ming Ho, and Der-Tsai Lee.
\newblock Travel-time prediction with support vector regression.
\newblock {\em IEEE Trans. Intell. Transp. Syst.}, 5(4):276--281, 2004.

\bibitem{yang2017ensemble}
Senyan Yang, Jianping Wu, Yiman Du, Yingqi He, and Xu~Chen.
\newblock Ensemble learning for short-term traffic prediction based on gradient
  boosting machine.
\newblock {\em J. Sensor.}, 2017, 2017.

\bibitem{yao2019revisiting}
Huaxiu Yao, Xianfeng Tang, Hua Wei, Guanjie Zheng, and Zhenhui Li.
\newblock Revisiting spatial-temporal similarity: A deep learning framework for
  traffic prediction.
\newblock In {\em Thirty-Third AAAI Conf. Artif. Intell. (AAAI)}, pages
  5669--5675, 2019.

\bibitem{yao2018deep}
Huaxiu Yao, Fei Wu, Jintao Ke, Xianfeng Tang, Yitian Jia, Siyu Lu, Pinghua
  Gong, Jieping Ye, and Zhenhui Li.
\newblock Deep multi-view spatial-temporal network for taxi demand prediction.
\newblock In {\em Thirty-Second AAAI Conf. Artif. Intell. (AAAI)}, 2018.

\bibitem{ye2012short}
Qing Ye, Wai~Yuen Szeto, and Sze~Chun Wong.
\newblock Short-term traffic speed forecasting based on data recorded at
  irregular intervals.
\newblock {\em IEEE Trans. Intell. Transp. Syst.}, 13(4):1727--1737, 2012.

\bibitem{yu2018spatio}
Bing Yu, Haoteng Yin, and Zhanxing Zhu.
\newblock Spatio-temporal graph convolutional networks: A deep learning
  framework for traffic forecasting.
\newblock In {\em Twenty-Seventh Int. Joint Conf. Artif. Intell. (IJCAI)},
  pages 3634--3640, 2018.

\bibitem{yu2017deep}
Rose Yu, Yaguang Li, Cyrus Shahabi, Ugur Demiryurek, and Yan Liu.
\newblock Deep learning: A generic approach for extreme condition traffic
  forecasting.
\newblock In {\em 2017 SIAM Int. Conf. Data Min. (SDM)}, pages 777--785. SIAM,
  2017.

\bibitem{zhang2017deep}
Junbo Zhang, Yu~Zheng, and Dekang Qi.
\newblock Deep spatio-temporal residual networks for citywide crowd flows
  prediction.
\newblock In {\em Thirty-First AAAI Conf. Artif. Intell. (AAAI)}, 2017.

\bibitem{zhang2021traffic}
Xiyue Zhang, Chao Huang, Yong Xu, Lianghao Xia, Peng Dai, Liefeng Bo, Junbo
  Zhang, and Yu~Zheng.
\newblock Traffic flow forecasting with spatial-temporal graph diffusion
  network.
\newblock In {\em AAAI Conf. Artif. Intell.}, volume~35, pages 15008--15015,
  2021.

\end{thebibliography}

\end{document}